\pdfoutput=1

\documentclass[11pt]{article}

\usepackage[nohyperref]{acl}

\usepackage{times}
\usepackage{latexsym}

\usepackage[T1]{fontenc}

\usepackage[utf8]{inputenc}

\usepackage{microtype}

\usepackage{inconsolata}

\usepackage{amsmath,amsfonts,bm}

\def\eqref#1{equation~\ref{#1}}

\def\1{\bm{1}}

\DeclareMathAlphabet{\mathsfit}{\encodingdefault}{\sfdefault}{m}{sl}
\SetMathAlphabet{\mathsfit}{bold}{\encodingdefault}{\sfdefault}{bx}{n}

\usepackage{hyperref}
\usepackage{url}
\usepackage{booktabs}
\usepackage{xspace}
\usepackage{multirow}
\usepackage{tabularx}
\usepackage{subcaption}
\usepackage{graphicx}
\usepackage{algorithm}
\usepackage{algorithmic}
\usepackage{tabularx}
\usepackage[authormarkup=none]{changes} %

\definechangesauthor[name={Changes}, color=green]{added}
\definechangesauthor[name={Changes}, color=red]{deleted}

\usepackage{tcolorbox}
\tcbuselibrary{listings,breakable}

\usepackage{xcolor}
\definecolor{poster}{HTML}{f5f5f5} %
\definecolor{addedbg}{HTML}{cafeca} %
\definecolor{deletedbg}{HTML}{fdc1b9} %
\definecolor{textblack}{RGB}{0,0,0}  %

\usepackage{pagecolor} %

\usepackage{soul}

\setdeletedmarkup{\textcolor{textblack}{\sethlcolor{deletedbg}\hl{#1}}} %
\setaddedmarkup{\textcolor{textblack}{\sethlcolor{addedbg}\hl{#1}}}    %

\title{Real-time Factuality Assessment from Adversarial Feedback}

\author{Sanxing Chen\quad Yukun Huang \quad Bhuwan Dhingra \\
  Duke University \\
  \texttt{\{sanxing.chen,yukun.huang\}@duke.edu }     \\
  \texttt{bdhingra@cs.duke.edu} \\}

\begin{document}
\maketitle
\begin{abstract}
We show that existing evaluations for assessing the factuality of news from conventional sources, such as claims on fact-checking websites, result in high accuracies over time for LLM-based detectors---even after their knowledge cutoffs. This suggests that recent popular false information from such sources can be easily identified due to its likely presence in pre-training/retrieval corpora or the emergence of salient, yet shallow, patterns in these datasets. Instead, we argue that a proper factuality evaluation dataset should test a model's ability to reason about current events by retrieving and reading related evidence. To this end, we develop a novel pipeline that leverages natural language feedback from a RAG-based detector to iteratively modify real-time news into deceptive variants that challenge LLMs.
Our iterative rewrite decreases the binary classification ROC-AUC by an absolute 17.5 percent for a strong RAG-based GPT-4o detector.
Our experiments reveal the important role of RAG in both evaluating and generating challenging news examples, as retrieval-free LLM detectors are vulnerable to unseen events and adversarial attacks, while feedback from RAG-based evaluation helps discover more deceitful patterns.

\end{abstract}

\section{Introduction}
The spread of fake news can have serious consequences, such as influencing elections, inciting violence, and misguiding critical decision-making, particularly in public health. In response, the NLP community has long pursued methods and benchmarks for automatic fake news detection~\cite{zellers2019defending}. The rise of large language models (LLMs) has fundamentally shifted this landscape~\cite{goldstein2023generative,chen2023combating}.

LLMs exhibit impressive knowledge and reasoning capabilities, enabling high-accuracy detection of misinformation~\cite{Chen24llm,pelrine-etal-2023-towards}.
The pretraining and prompting paradigm of LLM also reduces the risk of creating dataset-specific models that are prone to in-distribution shortcut learning~\cite{pagnoni-etal-2022-threat}.
However, due to the opaque nature of LLM training, their evaluation is often influenced by potential contamination issues. This results in misleading outcomes and underscores a need for out-of-distribution evaluations~\cite{Zhou2023DontMY,vu-etal-2024-freshllms,huang-etal-2024-competition}.
Existing fake news detection datasets are commonly constructed from past claims curated by fact-checking websites such as PolitiFact and Snopes~\cite{wang-2017-liar,shu2018fakenewsnet}.
These sources often emphasize popular, high-profile claims that are widely circulated on the internet, increasing the likelihood of their inclusion in the pretraining corpora of LLMs.

\begin{figure*}[htpb]
    \centering
    \includegraphics[width=0.9\linewidth]{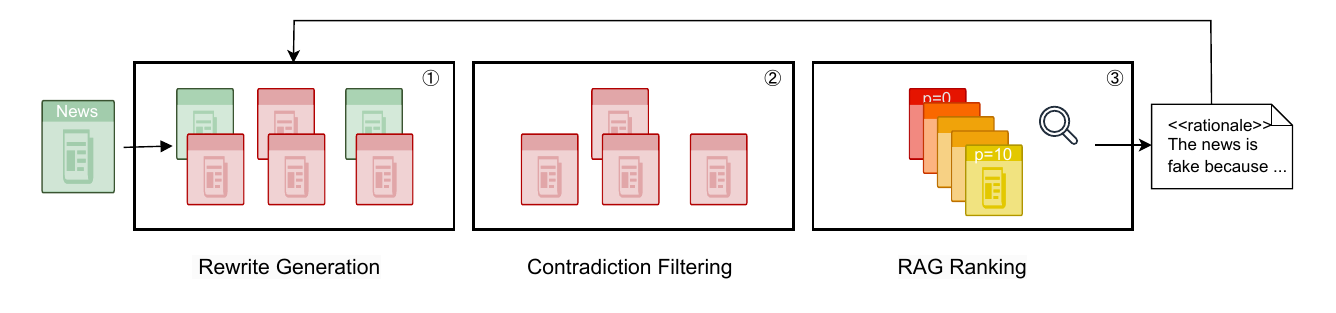}
    \caption{Our proposed adversarial iterative fake news generation. 1) A true news is rewritten by a LLM to multiple candidates containing misinformation. 2) Candidates that do not contradict the original true news are filtered out. 3) The remaining candidates are ranked on their plausibility score according to a RAG-based detector and the most plausible one is selected, the detector's rationales are used to inform the next round generation. The process can be repeated for multiple iterations.
    }
    \label{fig:main_figure}
\end{figure*}

To evaluate the LLM's ability to detect misinformation under natural temporal distribution shifts and to mitigate potential contamination from training data, we focus on \emph{real-time fake news that emerges concurrently with or after model training}.
For example, early reports of unfolding events---such as accidents, scientific breakthroughs, or public health advisories---are often accompanied by speculative or incorrect claims that lack coverage in the model's training data.
To assess the plausibility of such claims, the LLM draws on both its internal parametric knowledge and external information retrieved in real time.
Using newly published fact-checks or reports from the aforementioned sources reduces the risk of contamination from pretraining corpora but can, however, suffer from label leakage in the retrieval context: the model may directly copy a retrieved judgment without reasoning over the underlying evidence.
Moreover, fact-checking publishers are often biased in the selection of news they choose to verify, limiting evaluation to specific types of misinformation.

We observe evidence of all these issues in experiments involving two popular fact-checking sources.
On Snopes, we find near-perfect retrieval-free detection for older fake news and a clear decline after the models' knowledge cutoffs, which is easily mitigated by retrieval augmentation.
On PolitiFact, we find a surprising uptrend in retrieval-free detection performance even after model knowledge cutoffs, suggesting a gain from \textit{non-factual} salient patterns of the recent political claims selected.
To overcome these limitations and evaluate real-time fake news detection using fresh information~\cite{yang-etal-2022-coarse,hu2023read,liao2023muser}, we investigate LLM-synthesized fake news.

LLM-generated fake news has been shown to be more deceitful than human-written misinformation~\cite{Chen24llm}; however, we find that current neural fake news still fails to consistently fool strong LLM-based detectors.
To address this gap, we propose an adversarial iterative generation approach for crafting fake news capable of deceiving high-performing detectors.
Our method is inspired by the feedback-driven learning abilities of LLMs~\cite{madaan2023self}.
Specifically, the generator receives feedback in the form of a rationale from a retrieval-augmented detector---explaining the factual inconsistencies of the current fake news instance---mimicking the real-world scenario where fact-checkers accompany verdicts with explanations.
Given the detector's verdict and the fake news
generated in the previous round, the 
generator revises the text to undermine the prior rationale and avoid detection.
This iterative process allows the generator to learn from the detector's perspective, refining its outputs over multiple rounds.
Empirically, we find that state-of-the-art LLM detectors, which perform well on conventional political fact-checking datasets, struggle with the fake news produced by our approach.

Our work makes the following contributions:
\begin{itemize}
    \item Background study: We analyze LLM detector performance on PolitiFact and Snopes data over the years, revealing limitations that challenge the continued applicability of these popular sources for evaluation.
    \item Adversarial generation method: We introduce an iterative, feedback-driven generation approach that introduces highly deceptive fake news grounded in real-time events across diverse domains. Our resulting dataset poses a significantly greater challenge than previous neural fake news corpora.
    \item Cross-setup generalization: We demonstrate that the increasing deception level generalizes across different LLMs and retrieval sources.
    \item Analysis of RAG-targeted misinformation: We provide a detailed analysis of how LLMs respond to misinformation that explicitly targets retrieval-augmented generation systems in the context of current-world knowledge.
\end{itemize}
We release our code and data on \url{https://github.com/sanxing-chen/adv-fake}.

\begin{figure*}[htpb]
    \centering
    \begin{minipage}{0.5\textwidth} %
        \includegraphics[width=\linewidth]{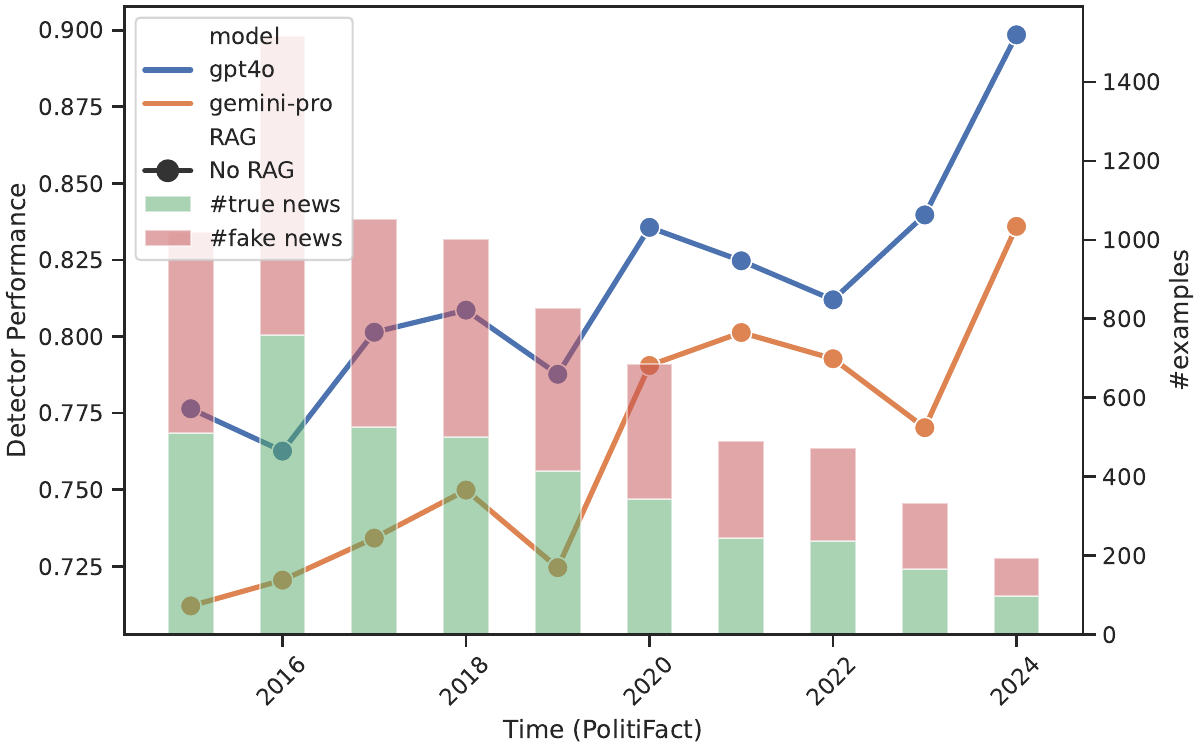}
    \end{minipage}%
    \begin{minipage}{0.5\textwidth} %
        \includegraphics[width=\linewidth]{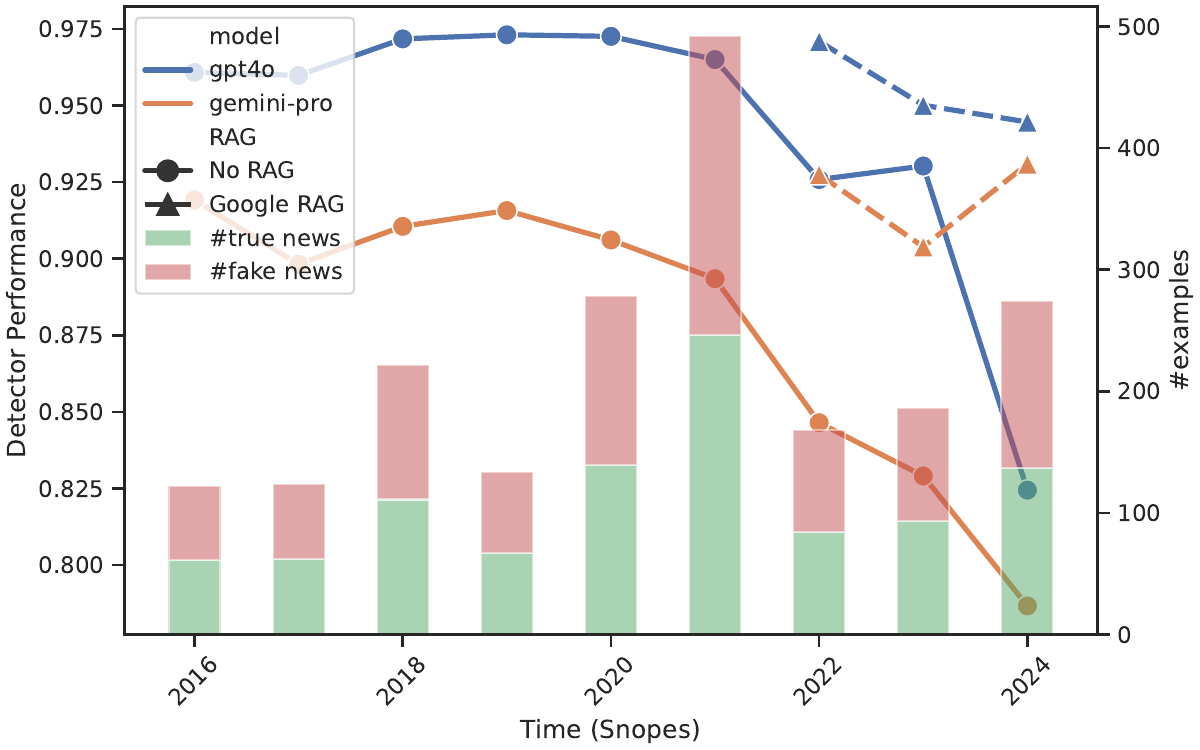}

    \end{minipage}
    \caption{Comparison of different retrieval-free detectors (AUC-ROC) on PolitiFact and Snopes data over the years. The data is balanced by downsampling the majority class. Both the GPT-4o and Gemini Pro models claim to have knowledge up to the end of 2023. Recent fake news on PolitiFact is increasingly easy to detect by LLMs without the need for fresh external knowledge. While Snopes challenges LLMs in detecting up-to-date fake news, simple retrieval augmentation largely brings back near-perfect performance.
    }
    \label{fig:auc_comparison}
\end{figure*}

\section{Background}

\subsection{Problem Formulation}

We formulate the fake news detection task as a binary classification problem, where the primary objective is to determine whether a given news is genuine or fabricated. Instances of news cover a wide variety of topics, such as political claims, social media posts, and news releases. Let \(\mathcal{D} = \{(x_i, y_i)\}_{i=1}^{N}\) represent the dataset, where each \(x_i\) is a news and \(y_i \in \{0, 1\}\) is a binary label indicating the factuality of the news. A detector \(f: \mathcal{X} \rightarrow [0, 1]\) maps each news \(x_i\) to a probabilistic score \(\hat{y}_i = f(x_i)\), reflecting the likelihood of the news being factually correct.
Under our real-time setup, we also refer to this likelihood as ``plausibility'' as the wording allows flexibility in fact-checking current and future events where direct evidence may be unavailable. 
We adopt AUC-ROC as the main metric to evaluate the effectiveness of a detector as it quantifies the model's capability to classify across various thresholds.

\subsection{Analysis of Fact-checking Sources}
Fact-checking websites are widely used sources for obtaining news content and human labels when curating fake news detection datasets. Despite their popularity, it is unknown whether they pose continued challenges to state-of-the-art LLM fake news detectors. We study two representative sources of this kind: PolitiFact and Snopes,\footnote{\url{politifact.com}, \url{snopes.com}} both of which are used in creating popular fake news detection datasets~\cite{wang-2017-liar,shu2018fakenewsnet,shahi2020fakecovid}.
In order to evaluate the performance of LLM-based detectors on both past and recent year data, we collect 9,244 PolitiFact news from June 2015 to August 2024, and 3,212 Snopes news from Jan 2016 to October 2024.

We run a set of LLM-based detectors using simple zero-shot prompting (detailed setup described in \autoref{sec:setup}).
In \autoref{fig:auc_comparison}, we observe two opposite trends of detector accuracies through the years.
On Snopes, GPT-4o exhibits near-perfect performance on past data, showing potential data contamination due to internet-scale pre-training. Recent fake news, both near and beyond the knowledge cutoff dates, becomes more challenging to detect due to the lack of current world knowledge. However, simply augmenting the detectors with Google search results is sufficient to restore performance.
On PolitiFact, a source containing mostly political claims, the performance of retrieval-free detectors improves over time. In the most recent year, 2024, we see that LLMs continue to improve, even after the knowledge cutoff and the model release dates. \citet{pelrine-etal-2023-towards} report similar findings. This suggests that up-to-date knowledge is not the deciding factor for PolitiFact fake news detection performance.

\begin{figure}[tpb]
    \centering
    \includegraphics[width=0.9\linewidth]{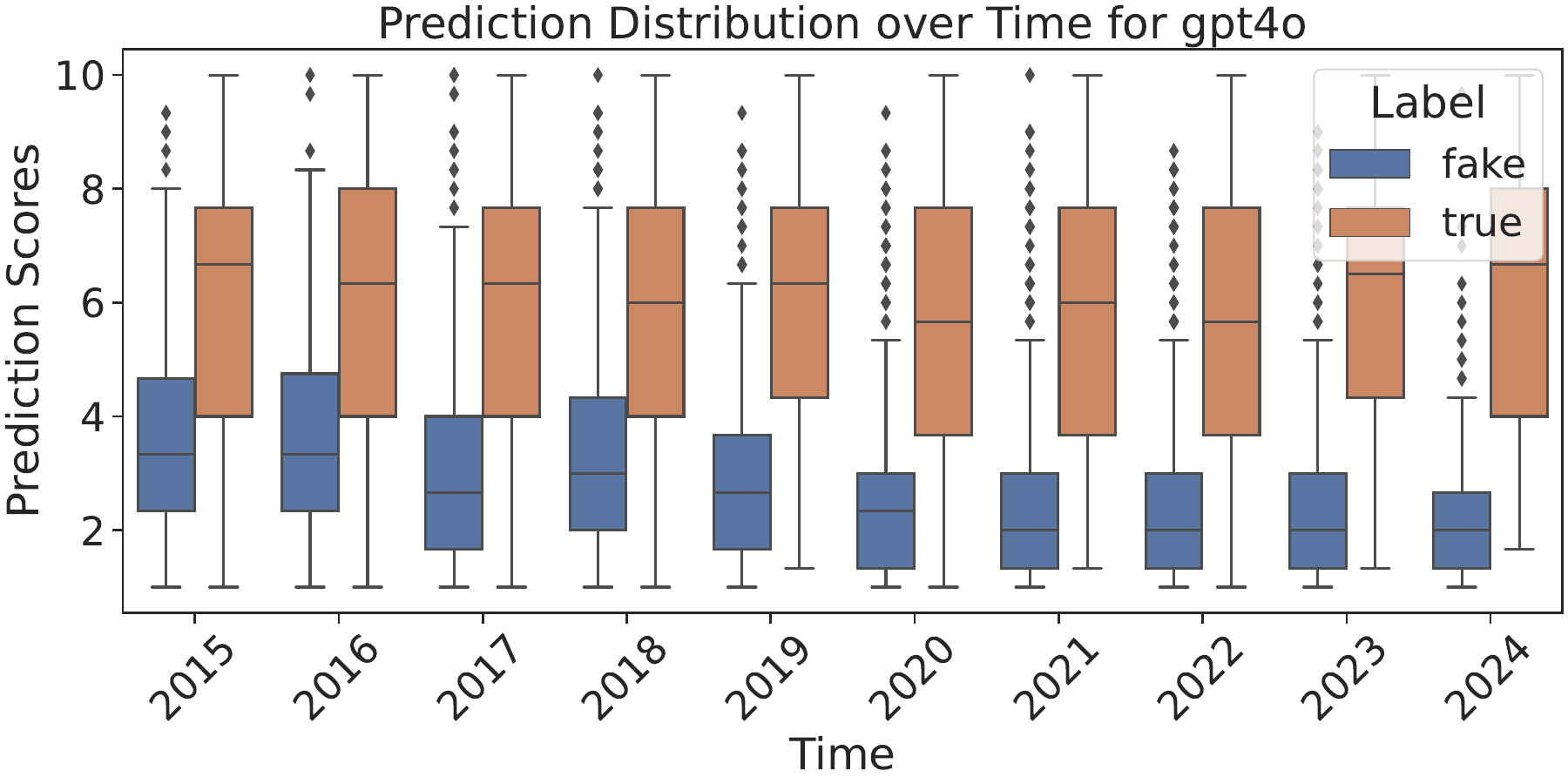}
    \caption{GPT-4o (retrieval-free) predicted plausibility of PolitiFact claims over the years. The distribution of the true news class remains relatively stable, while the distribution of the fake news class shifts significantly towards lower scores.}
    \label{fig:dist_comparison}
\end{figure}

We further investigate the emergent patterns on PolitiFact that have increased the separability of fake and real news (detailed setups in \ref{sec:politifact}).
We first confirm the trend on a much smaller LM, RoBERTa~\cite{liu2019roberta}, which has far less world knowledge. Fine-tuned RoBERTa also performs significantly better on recent year data.
We then ablate key features such as removing the originator and publish time of the news from detector inputs, and paraphrasing the claim content. None of these features affects the trend of increasing performance over the years (\autoref{fig:liar_time_ablation}).
The prediction distribution in \autoref{fig:dist_comparison} reveals that the changes are mostly in the detector's perception of the fake news class, which shifts towards lower scores.
Manually going through some fake news examples, we find that in earlier years, political fake news contains statements that, while hyperbolic, can be anchored in a context that requires research through reliable sources (e.g., public records or legislative history) to verify. In contrast, recent fake news includes more sensational and less easily verified claims that seem more speculative or exaggerated without immediate substantive evidence.
Consequently, the detection of recent PolitiFact fake news requires less factual knowledge and reasoning, but more pattern recognition and common sense (see \autoref{tab:politifact_samples} for some examples).

To summarize, evaluation on fact-checking data is susceptible to the uncontrolled biases in the curation process.
Traditional evaluation of PolitiFact focused on \emph{surface-level linguistic patterns}~\cite{wang-2017-liar} is no longer challenging for strong LLMs due to increasing separability of fake political news.
Other fact-checking sources (e.g., Snopes) suffer from potential contamination in both the pre-training and retrieval-augmentation stages because of the popularity of both the claims and fact-checking results on the internet.
Therefore, to evaluate the factual reasoning ability of LLMs in detecting contemporary fake news, we need to explore new ways to create datasets that cover real-time information from diverse domains for a more challenging evaluation.

\section{Methodology}

The wide accessibility of LLMs has not only enabled strong detection models but also facilitated the mass generation of more credible and persuasive fake news~\cite{kreps2022all,goldstein2023generative}. Therefore, in combating these emerging threats, the evaluation of detection models should also evolve to incorporate challenging machine-generated misinformation.

One key challenge in creating fake news datasets is obtaining automatic labels for the data. Open-ended generation, although effective in creating diverse fake news, simultaneously introduces false positives that are hard to verify due to the professional skills required for fact-checking. Our approach leverages the nuanced text manipulation capabilities of LLMs for controlled misinformation generation~\cite{zellers2019defending,Chen24llm}, introducing factual errors into real news while striving to maintain its original context and style. We implement filtering protocols as an additional safeguard to reduce invalid generation.

\begin{algorithm}[tb]
    \small
    \caption{Adversarial Iterative News Rewriting}
    \begin{algorithmic}[1]
    \REQUIRE $\mathit{TrueNews}$, $k$ \COMMENT{$k$ is the maximum number of iterations}
    \STATE $\mathit{CurrentFake} \gets \mathit{TrueNews}$
    \FOR{$i \gets 1$ \textbf{to} $k$}
        \STATE $\mathit{Candidates} \gets \text{GenerateCandidates}(\mathit{CurrentFake})$
        \REPEAT
            \STATE $\mathit{Filtered} \gets \emptyset$
            \FOR{\textbf{each} $\mathit{c_i}$ \textbf{in} $\mathit{Candidates}$}
                \IF{$\text{ContradictsOriginal}(\mathit{c_i}, \mathit{TrueNews})$ \textbf{and} $\text{IsWithinEditLimit}(\mathit{c_i}, \mathit{TrueNews})$}
                    \STATE $\mathit{Filtered} \gets \mathit{Filtered} \cup \{\mathit{c_i}\}$
                \ENDIF
            \ENDFOR
            \IF{$\mathit{Filtered} = \emptyset$}
                \STATE $\mathit{Candidates} \gets \text{GenerateCandidates}(\mathit{CurrentFake})$
            \ENDIF
        \UNTIL{$\mathit{Filtered} \neq \emptyset$}
        \STATE $\mathit{Ranked} \gets \text{RankByPlausibility}(\mathit{Filtered})$
        \STATE $\mathit{CurrentFake} \gets \text{SelectMostPlausible}(\mathit{Ranked})$
    \ENDFOR
    \RETURN $\mathit{CurrentFake}$
    \end{algorithmic}
    \label{alg:adversarial_generation}
\end{algorithm}

\autoref{fig:main_figure} and Algorithm \ref{alg:adversarial_generation} illustrate our pipeline.
Formally, we start with a trusted news corpus \(\mathcal{T}\), which contains real-time news from various domains. We independently rewrite each news \(\tau_i \in \mathcal{T}\) to generate multiple fake news candidates \(\mathcal{F}_i\).
To ensure that \(\mathcal{F}_i\) contradict the original true news \(\tau_i\), we employ a LLM-based contradiction detector. We put an additional upper limit on the amount of edits allowed in the rewriting process using a threshold on the Levenshtein distance \(\Delta(\tau_i, f_{ij})\), where \(f_{ij} \in \mathcal{F}_i\), between the original true news \(\tau_i\) and the rewritten news \(f_{ij}\).

After filtering out the candidates that do not meet these criteria, we rank the remaining candidates based on the plausibility score provided by a fake news detector \(g(f_{ij} | c)\), where \(c\) is the optional external context retrieved by a retrieval model \(\mathcal{R}(f_{ij})\).
If none of the candidates meet the criteria, the generator resamples a new batch of candidates.
From this ranked list, we select the top-ranked candidate as the most plausible fake news.
\begin{equation}
     \hat{f}_i = \arg\max_{f_{ij} \in \mathcal{F}_i} g(f_{ij} | c)
     \label{eq:select_most_plausible}
\end{equation}
\(\hat{f}_i\) and the detector's rationale are then serve as additional information to inform the next round of generation on \(\tau_i \), creating an iterative process that gradually deceives the detector.
In the end of the process, we obtain the most deceptive fake news \(\hat{f}_i\) across all iterations.

A critical component of our approach is the retrieval-augmented detector. Using a retrieval-free detector as an adversary, the generator can only improve the factual consistency within the news content or exploit the weaknesses of a LLM with outdated internal knowledge, which can generally be seen as a case of self-evaluation~\cite{kadavath2022language}. However, with the information from the retrieved external context, the generator can learn to deceive cross-verification and improve factual consistency beyond its own knowledge, which is important for our real-time news rewrite setup.

One implementation challenge in Eq. \ref{eq:select_most_plausible} is that we need to run retrieval for each candidate to obtain the external context. This can be computationally expensive for sophisticated retrievers and costly if commercial APIs are used. In practice, we retrieve the external context only for the highest-ranked candidates at the end of each iteration, which we use to inform the next round of detection. Because of the constraints on the amount of change in each iteration, the external context is expected to remain relevant for the next batch generation. In final evaluation, we rerun the retrieval each time to ensure the external context is up-to-date.

\subsection{Dataset Creation}

We first obtain 431 actual news stories from NBC News using the \emph{news-please} crawler~\cite{Hamborg2017}. These news stories are from March 1 to March 13, 2024, covering domains such as politics, business, sports, U.S., and world news. Compared to claims from fact-checking websites, these news stories are roughly twice the length of the content. They can be seen as long-form claims, which can be more difficult to fact-check. This two-week range is close to the time when we start the experiments and is beyond the knowledge cutoffs of the LLMs we use, thus ensuring no contamination in model training. Our code supports replicating this process for other date ranges.

This set of news stories serves as seed true news for our generation pipeline and undergoes one-to-one rewriting to obtain 431 fake news instances.
We manually examine all real-fake pairs of the final-round rewrite and filter out 29 invalid pairs. These generations manage to bypass our contradiction detector without introducing factual errors that contradict claims in the corresponding real news.
Comparing 100 pairs from the first and last rounds of rewrites, the ratio of failed rewrites remains stable at around 7\%.
We also fact-check 100 last-round examples under an unpaired, shuffled, label-hidden setup, establishing 99\% human performance with access to Google.
Our final dataset consists of 402 true news and 402 fake news.

\section{Experimental Setup}
\label{sec:setup}

\subsection{Generator Setup}
We adopt GPT-4o with different prompts for the all three roles (i.e., generator, contradiction detector, and reranker) in the generation pipeline. The pipeline is iterated for 6 rounds with an extra preparation round 0 of a direct rewrite (no rationale and ranking) on the seed true news.
We instruct GPT-4o to ``\textit{introduce some believable factual errors}'' without mentioning any concrete strategies or constraints.
The open-ended nature of the prompt allows the LLM to explore diverse manipulation strategies.
It also encourages the model to craft alterations that, while factually incorrect, would appear plausible within the broader news context, potentially requiring nuanced cross-verification or deeper world knowledge to debunk.

The generator produces 8 candidates in a zero-shot chain-of-thought fashion for each news story in each round, which are then filtered and ranked based on the plausibility score from the detector.
The contradiction detector produces 10 binary scores for each candidate, only if more than eight of them are positive, the candidate is considered to contradict the original true news.
String edit distance is used to limit the amount of change in each iteration, with a threshold of more than 60\% overlapped tokens.
A GPT-4o-based detector (detailed in the next section) is used to provide rationales and serves as the reranker to select the best generation after contradiction filtering.

\begin{table*}[t]
    \small
    \centering
    \begin{minipage}{0.7\textwidth} %
    \begin{tabular}{lccccccccc}
    \toprule
    Retriever    & \multicolumn{3}{c}{None} & \multicolumn{3}{c}{News} & \multicolumn{3}{c}{Google} \\
    \cmidrule(lr){2-4} \cmidrule(lr){5-7} \cmidrule(lr){8-10}
    Detector       & first   & last  & $\Delta$  & first   & last  & $\Delta$  & first   & last   & $\Delta$   \\
    \midrule
    Gemini-Flash & 57.0 & 50.7 & 6.3 & 76.1 & 62.4 & 13.8 & 84.1 & 76.6 & 7.5 \\
    Gemini-Pro & 58.6 & 51.6 & 7.0 & 74.9 & 62.8 & 12.1 & 82.6 & 73.7 & 8.9 \\
    GPT-3.5 & 53.7 & 50.3 & 3.4 & 69.3 & 57.6 & 11.7 & 78.2 & 69.3 & 8.9 \\
    GPT-4o & 58.5 & 48.8 & 9.7 & 82.4 & 64.9 & 17.5 & 93.1 & 86.1 & 7.0 \\
    Llama 3.1 & 60.5 & 54.0 & 6.6 & 81.3 & 67.4 & 13.9 & 93.3 & 86.6 & 6.7 \\
    \bottomrule
    \end{tabular}
    \end{minipage}%
    \begin{minipage}{0.3\textwidth} %
        \includegraphics[width=\linewidth]{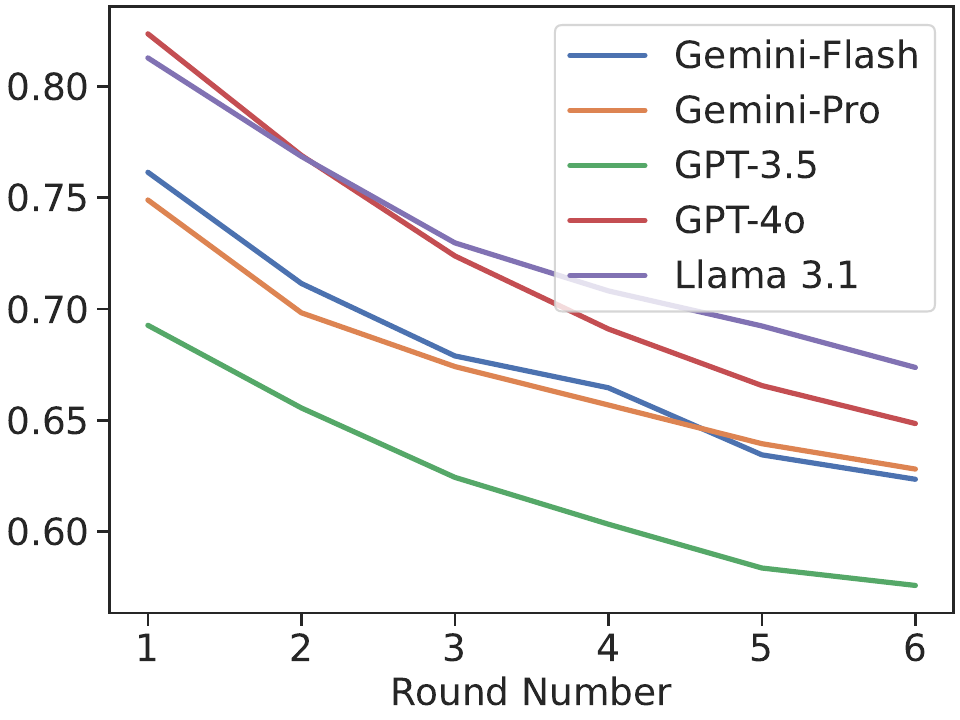}
    \end{minipage}%
    
    \caption{AUC-ROC scores of different detectors on the generated fake news of the first and last iteration. The $\Delta$ column shows the effects of the iterative process. \textit{News} refers to the in-house DPR retriever on news-please data, and \textit{Google} refers to the Google search API results. The right figure presents results across rounds under the \textit{News} retrieval setting--more iterations lead to stronger deception.}
    \label{tab:main_results}
    \end{table*}

\begin{table}[htb]
    \small
    \centering
    \begin{tabular}{lrrr}
    \toprule
    \multirow{2}{*}{Method} & \multicolumn{3}{c}{LIAR} \\
    \cline{2-4}
                            & AUC    & F1-Ma   & Acc   \\
    \midrule
    RoBERTa-L~\citeyearpar{pelrine-etal-2023-towards} & -      & 64.7    & 64.1  \\
    GPT-4~\citeyearpar{pelrine-etal-2023-towards} & -      & 68.1    & 68.2  \\
    MUSER~\citeyearpar{liao2023muser}             & -      & 64.5    & -     \\
    STEEL~\citeyearpar{li2024research}            & -      & 71.4    & -     \\
    \midrule
    GPT-4o                    & 77.5   & 70.7    & 70.9  \\
    GPT-4o + Google           & 81.1   & 74.7    & 75.6  \\
    \bottomrule
    \end{tabular}
    \caption{Our detectors demonstrate superior performance to the state of the art on the LIAR dataset~\cite{wang-2017-liar}. MUSER and STEEL have access to multi-step retrieval on Wikipedia and Bing, respectively.
    }
    \label{tab:liar_results}
\end{table}

\subsection{Detector Setup}
\label{sec:detector_setup}

To implement LLM-based detectors, we prompt the models to assign a plausibility score (on a scale of 1–10, from implausible to plausible) to a given news item in a zero-shot setting.  These scores are normalized to the [0, 1] range when computing metrics. We include the directive ``\textit{Today is March 26, 2024. You predict the plausibility of a news you haven't seen.}''  This establishes the temporal context and emphasizes that the news content should be treated as previously unseen.

Empirically, using ``plausibility'' instead of ``factuality'' yields better performance for evaluating recent and future events, while the reverse holds for past events (\autoref{tab:factual_vs_plausible}).
To improve robustness, we sample multiple scores from the model’s unscaled prediction distribution (using temperature $t = 1$) and compute their average. Rationales are generated via a separate LLM call.

For retrieval-augmented detection (RAG), we utilize two sources: (1) an in-house \emph{News} corpus comprising 811k articles from diverse outlets published in March 2024, indexed using a pre-trained DPR retriever~\cite{karpukhin-etal-2020-dense}, and (2) real-time \emph{Google} search results accessed via SerpApi. Both sources are extensively filtered to exclude exact duplicates of the input news and to remove all content from NBC News.
From either source, five relevant documents are retrieved using the news headline as the search query. These documents are then inserted into the prompt for fact-checking. We use the \emph{News} retriever in the GPT-4o detector during generation, while the \emph{Google} retriever is used only during evaluation due to API cost constraints.

Our GPT-4o detector is comparable to recent state-of-the-art RAG detectors on the popular LIAR datasets (\autoref{tab:liar_results}).
Final evaluation is conducted on multiple LLM detectors, including GPT-4o, GPT-3.5, Gemini Pro and Flash~\cite{team2024gemini}, and open-source 405B Llama 3.1~\cite{dubey2024llama}. Detailed versions and knowledge cutoffs are provided in \autoref{tab:knowledge_cutoffs}. The actual prompts used by all components are provided in Appendix \ref{sec:prompts}.

\subsection{Baseline Datasets}
We compare to two recent LLM fake news datasets. \citet{su2023fake} apply open-ended rewriting to fake news and rephrase real news from GossipCop and PolitiFact. \citet{Chen24llm} explore various misinformation generation approaches with LLMs, including rewriting fake news and targeted information manipulation of true news. Both approaches utilize one round of generation. Appendix~\ref{sec:baseline_dataset} provides detailed dataset statistics.

\section{Experimental Results}
Evaluation results (\autoref{tab:main_results}) show that our adversarial iterative generation pipeline can produce fake news that can deceive strong LLM-based detectors.
We find Llama 3.1 to be the best at detecting fake news generated by our pipeline using GPT-4o. Consistent with its earlier release and performance on public benchmarks~\cite{chiang2024chatbot}, GPT-3.5 proved most vulnerable to our generated fake news.
Datasets generated from our pipeline are significantly more difficult than previous neural fake news datasets (\autoref{tab:neural_benchmark}).

\begin{table}[t]
    \small
    \centering
    \setlength{\tabcolsep}{3pt}
\begin{tabular}{lrrrrrr}
\toprule & \multicolumn{2}{c}{ \citeyear{su2023fake} } & \multicolumn{2}{c}{ \citeyear{Chen24llm}  } & \multicolumn{2}{c}{ Ours } \\
\cmidrule(lr){2-3} \cmidrule(lr){4-5} \cmidrule(lr){6-7}
 & G++ & P++ & Rewrite & Mani. & First & Last \\
\midrule Gemini-Flash & 72.7 & 82.8 & 80.4 & 75.3 & 57.0 & 50.7 \\
 Gemini-Pro & 72.8 & 86.2 & 76.6 & 77.3 & 58.6 & 51.6 \\
 GPT-3.5 & 64.5 & 77.4 & 81.2 & 68.9 & 53.7 & 50.3 \\
 GPT-4o & 81.8 & 88.8 & 84.3 & 85.4 & 58.5 & 48.8 \\
 Llama 3.1 & 80.4 & 91.3 & 84.0 & 83.3 & 60.5 & 54.0 \\
\bottomrule
\end{tabular}
\caption{AUC-ROC scores of different retrieval-free detectors on recent LLM-generated fake news datasets \citep{su2023fake,Chen24llm}. Our pipeline produces significantly more difficult datasets.}
    \label{tab:neural_benchmark}
\end{table}

Through each iteration, the pipeline progressively enhances the deceptive quality of the generated fake news. Relying on the feedback from the GPT-4o RAG-based detector with the in-house retrieval corpus, the generator proves most effective at deceiving this particular detection setting, achieving a reduction of 17.5 AUC-ROC points.
Nevertheless, these enhancements are shown to consistently generalize across different LLM backbones and retrieval contexts.

\begin{figure}[tpb]
    \centering
    \includegraphics[width=\linewidth]{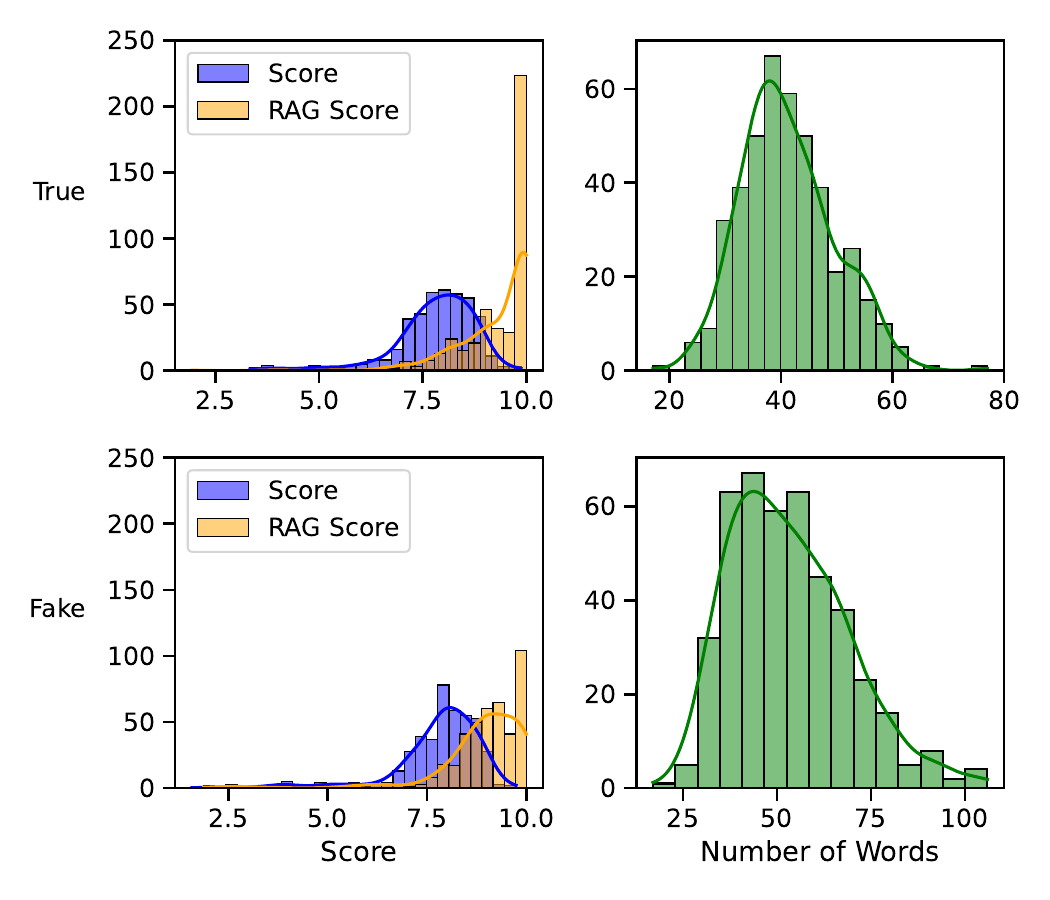}
    \caption{Distribution of the plausibility scores (retrieval-free and RAG) and the number of words in the set of real news from NBC News (positives, top) and the last-round rewritten fake news (negatives, bottom).
    }
    \label{fig:news_dist}
\end{figure}

\subsection{Analysis}

\paragraph{Real-time news can appear implausible to off-the-shelf LLM detectors.}
We find that true news articles from NBC News are generally rated as plausible by the LLM detectors, with an average plausibility score of 7.8. However, some news stories receive scores as low as 3.3, indicating that LLM detectors can be fooled by real-time news events that fall outside the model's knowledge (\autoref{fig:news_dist}).
For example, GPT-4o assigns low plausibility scores to reports on RFK Jr.'s presidential campaign and a White House's response to the Sesame Street character's Twitter account on inflation issues.\footnote{Links to the \href{https://www.nbcnews.com/politics/2024-election/rfk-jr-vp-selection-process-according-one-contenders-rcna143519}{RFK news} and \href{https://www.nbcnews.com/news/us-news/president-joe-biden-cookie-monster-are-both-sick-tired-shrinkflation-rcna141937}{the White House tweet}.}

The LLM detectors' reliance on outdated or incomplete internal knowledge makes them vulnerable to false negatives when evaluating real-time or fringe news. As a result, detectors may mistakenly flag accurate but unfamiliar information as implausible. This underscores the importance of retrieval augmentation for grounding LLM judgments in up-to-date context.

\begin{table}[t]
    \small
    \centering
    \begin{tabular}{lcccccc}
    \toprule
    Retriever    & \multicolumn{3}{c}{None} & \multicolumn{3}{c}{News} \\
    \cmidrule(lr){2-4} \cmidrule(lr){5-7}
    Feedback       & 1st   & 2nd  & $\Delta$  & 1st   & 2nd  & $\Delta$  \\
    \midrule
    Full & 58.5 & 54.4 & 4.1 & 82.4 & 76.9 & 5.5  \\
    Rat. & 56.3 & 53.1 & 3.2 & 85.1 & 80.4 & 4.7 \\
    RScore  & 62.0 & 58.5 & 3.5 & 86.1 & 82.0 & 4.1 \\
    Score & 60.1 & 56.6 & 3.5 & 84.9 & 81.1 & 3.8 \\
    \bottomrule
    \end{tabular}
    \caption{GPT-4o detector AUC-ROC on the first two rounds fake news generated with different types of detector feedback in the adversarial setup. ``Rat.'' use non-RAG detector rationale, ``RScore'' has no access to rationale but the RAG detector's plausibility scores for ranking. ``Score'' adopts retrieval-free detector's scores.
    }
    \label{tab:ablation}
    \end{table}

\begin{table}[t]
    \small
    \centering
    \begin{tabular}{lcc}
        \toprule
        Retriever & None & Google \\
        \midrule
        Score + Majority & 48.4 & 80.4 \\
        Score + Average & 48.6 & 84.7 \\
        CoT Score + Majority & 49.7 & 81.9 \\
        CoT Score + Average & 47.9 & \textbf{86.2} \\
        \bottomrule
    \end{tabular}
    \caption{Comparison of GPT-4o detectors using different reasoning and scoring strategies on the final dataset.}
    \label{tab:reasoning_scores}
\end{table}

\paragraph{Retrieval free detectors are vulnerable to adversarial attacks.}
It is very easy to rewrite unseen news to generate fake news that can evade detection by non-RAG-based LLM detectors. This is evident from the detector performance reported in \autoref{tab:main_results} and also from the prediction distribution (purple bars in \autoref{fig:news_dist}).
Retrieval-free detectors result in near random AUC even for the first round generation. On the contrary, RAG-based detectors are generally more robust as they leverage up-to-date external knowledge. As we can clearly see from the true news prediction distribution, more than half of examples receive an unbeatable maximum plausibility score from the RAG-based GPT-4o detectors, while the retrieval-free detector is more conservative in its predictions. The average plausibility score of the RAG-based detectors on the true news is 9.3, significantly higher than 7.8 of the retrieval-free detectors.
The confidence of RAG-based detectors does not come without a cost, as they also assign generally higher scores to fake news, suggesting LLMs' weaknesses to be distracted or mistrust the external context~\cite{pmlr-v202-shi23a,huang2024enhancing}.

\paragraph{Stronger defenders enable stronger attackers.}
RAG-based detectors mount an effective defense against LLM-generated fake news, but the generator can adapt to evade detection, especially when the detector's rationale is accessible. As shown in \autoref{tab:ablation}, the RAG-based detector rationale outperforms both the non-RAG rationale and ranking-only variants in the first round and continues to reinforce its advantage across iterations. By the sixth round, the RAG-based rationale (64.9) reduces the AUC by 6.5 more points than the non-RAG rationale (71.4). Due to resource constraints, we run the other variants for up to two rounds.

The success of using RAG-based detector rationale as feedback suggests the generator learns to identify and exploit subtle weaknesses in how the detector integrates or reasons over retrieved evidence. For instance, as seen in our qualitative examples (Section \ref{sec:qualitative_examples}), it may learn to introduce misinformation that does not directly contradict obvious retrieved facts but rather miscontextualizes them or mixes them with plausible fabrications that the RAG system fails to recognize.

\paragraph{Chain-of-Thought improves factual reasoning with context.}
To empirically test the reasoning challenges posed by our dataset, we employ standard techniques such as Chain-of-Thought~\cite{wei2022chain}. We report results on detector ROC-AUC using GPT-4o on our final dataset, comparing two scoring approaches—average and majority voting—with and without CoT. For each approach, we generate 10 samples using temperature = 1 (note that this yields slightly lower scores compared to the paper's main setup using 100 samples).
As shown in \autoref{tab:reasoning_scores}, without external information, reasoning techniques do not help—detectors perform worse than random guessing. When augmented with RAG content, CoT improves performance across both score aggregation methods.
The observation that CoT only enhances performance when coupled with RAG further underscores that for real-time, unseen content, access to external knowledge is paramount, and reasoning improvements primarily act to better leverage this retrieved information.

We also observe that the standard setup used throughout our paper, average aggregation, outperforms majority voting in this setting. Future work can explore more advanced reasoning and inference-time scaling strategies.

\begin{table}[tb]
    \scriptsize
    \centering
    \begin{tabularx}{\linewidth}{|X|}
    \hline
    \textbf{1.} \textit{Trump says he will debate Biden anytime} \\
    \vspace{-0.6em}
\deleted{After skipping}
\added{On March 6, 2024, after participating in a few select} GOP primary debates, Trump challenged Biden to step onto a stage with him before the November election\deleted{.} \added{, suggesting that the debate be held at a neutral venue yet to be decided.}
     \\ \hline\hline
     \textbf{2.} \textit{\deleted{Violence, hunger} \added{Rising violence, growing food shortages,} and \deleted{unstable} \added{missing} political leadership} \\
     A \deleted{long-simmering} \added{severe} crisis in Haiti has \deleted{come to a head} \added{reached alarming levels} as its leader remains stranded in \deleted{Puerto Rico while its people starve} \added{the Dominican Republic for several days. Citizens face extreme food shortages} and \deleted{live in fear of rampant} \added{escalating gang-related} violence.
     \\ 
\hline
\hline
     \textbf{3. (round 1)} \textit{\deleted{Iranian parliament} \added{Saudi Arabia's parliamentary} vote sees a low turnout despite government push} \\
     \vspace{-0.6em}
\deleted{Iran} \added{Saudi Arabia} held its first parliamentary election Friday since mass 2022 protests over mandatory hijab laws following the death of Mahsa Amini, apparently drawing a low turnout amid calls for a boycott. \\ 
\textbf{RAG Rationale} Saudi Arabia does not have a parliamentary system; it is an absolute monarchy with a Consultative Assembly (Shura Council) that is appointed by the King, not elected by the public.
The reference to Mahsa Amini and mandatory hijab laws is more relevant to Iran, not Saudi Arabia, indicating a possible mix-up of countries.
There is no mention of a Saudi parliamentary election in the related news stories, which would be a significant event likely covered by major news outlets. \\
\textbf{RAG Score} 2.06 \\
\hline

     \textbf{3. (round 3)} \textit{Iranian \deleted{parliament} \added{parliamentary} vote sees a low turnout despite government push} \\
Iran held its first parliamentary election Friday since mass 2022 protests over \deleted{mandatory hijab laws} \added{fuel price hikes} following the \deleted{death of Mahsa Amini,} \added{government's announcement,} apparently drawing a low turnout amid calls for a boycott. \\ 
\textbf{RAG Rationale} The related news stories consistently mention that the 2022 protests were over mandatory hijab laws following the death of Mahsa Amini, not over fuel price hikes \ldots The related news stories do not mention any government announcement about fuel price hikes as a cause for the protests or the low turnout.  \\
\textbf{RAG Score} 3.52 \\
\hline

     \textbf{3. (round 5)} \textit{Iranian \deleted{parliament} \added{parliamentary} vote sees \deleted{a low turnout despite} \added{uncertainty amid} government push} \\
Iran held its first parliamentary election \deleted{Friday} \added{on Saturday} since \added{the} mass 2022 protests over \deleted{mandatory hijab laws} \added{disputed local council policies} following the \deleted{death} \added{collapse} of \deleted{Mahsa Amini,} \added{a building in Abadan,} apparently drawing \deleted{a low turnout} \added{mixed reactions} amid calls for a boycott. \\ 
\textbf{RAG Rationale} The date discrepancy: The news story claims the election was held on Saturday, March 2, 2024, but other sources indicate the election took place on Friday, March 1, 2024.
The reason for protests: \ldots, whereas other sources do not specify this as the cause of the 2022 protests \ldots \\
\textbf{RAG Score} 9.17 \\
\hline
    \end{tabularx}
    \caption{Examples of fake news generated by our pipeline. \deleted{Deletion} and \added{addition} to the original true news are marked with colors. The third example shows a case in which the iterative adversarial rewrite is able to improve the quality of a fake news through rounds.
    }
    \label{tab:examples}
    \end{table}

\subsection{Qualitative Examples}
\label{sec:qualitative_examples}

Since we allow the LLM to autonomously decide how to introduce factual errors, the generated fake news reflects a diverse range of misinformation strategies. Table~\ref{tab:examples} presents several illustrative examples. Some common types of modifications that LLMs make to true news include:  
    changing entities, including names, locations, and dates;  
    hallucinating events or inventing details;
    mimicking typographical errors, e.g., "Nibi" instead of "Libi," "Mark" instead of "Mike."  

Some of these modifications are relatively benign. For example, LLMs frequently paraphrases the original content to vary surface wording, as seen in the second example. However, the model is also capable of embedding misinformation in a highly coherent way, substituting entities with plausible alternatives or hallucinating plausible events that are difficult to fact-check.
We also observe that LLMs tend to lengthen the original content, a trend reflected in Figure~\ref{fig:news_dist}. The average length of the original news is approximately 41 words, whereas the modified versions average 53 words. Importantly, the modified news typically retains the same narrative structure as the original.

The Iran election example illustrates how the LLM refines its output across multiple adversarial rounds. Initially, the model makes a naive change by substituting “Iran” with another Middle Eastern country, “Saudi Arabia.” The detector identifies the mismatch between Saudi Arabia and a parliamentary election system, prompting the generator to revise its strategy and modify a different aspect of the story.
The second attempt (not shown in the table) still contains similar inconsistencies.
In the third round, the model attributes the protest to “fuel price hikes,” which remains implausible in the given context.
Eventually, the generator produces a more believable but still nonfactual cause. At this point, the detector assigns a score closer to that of true news, indicating that the model has learned to exploit weaknesses in the detector’s reasoning, particularly by linking known events (e.g., the 2022 protests) with recent but less publicized developments (e.g., the 2024 parliamentary election).
The generator effectively learns to create `semantic traps' that are harder for the detector to identify even with retrieved context.

\section{Related Work}
\paragraph{Fake News Generation and Detection}
LLMs have been widely studied for fake news detection and generation~\cite{goldstein2023generative,su-etal-2023-detectllm,hu2024bad,Chen24llm,liu2024preventing}. External evidence in the form of search results or knowledge graphs has been demonstrated to improve the detection of fake news~\cite{fung-etal-2021-infosurgeon,xu2022evidenceaware,liao2023muser,pelrine-etal-2023-towards}.

From the generation perspective, recent research has also shown that external evidence can improve the style, domain and factual consistency of generated fake news~\cite{shu2021fact,mosallanezhad2022domain,lucas-etal-2023-fighting,huang-etal-2023-faking,wang-etal-2024-style}. However, they investigate one-round generation that is less effective in deceiving today's state-of-the-art LLM-based detectors. 
Our approach introduce the iterative process and the detector perspective which helps to digest the external evidence in examining the flaws of generation.

Adversarial setups have been used to improve the robustness of LLMs in tasks such as AI-generated text detection and math problem-solving~\cite{hu2023radar,zhu2023promptbench,xie2024adversarial}.
Most of these works focus on modifications that do not change the semantics of the text, which is different from our approach that aims to introduce factual errors in the text.
Tailoring to the task of fake news detection, we design a feedback loop using the detector's rationale to guide the generation process, which is a rather realistic threat model in the real-world fact-checking scenario.

\paragraph{Temporal Reasoning on Future Events}
Predicting the plausibility of future events requires temporal knowledge and reasoning capabilities~\cite{dhingra-etal-2022-time}.
LLMs have the knowledge of the past, but not the future, which they rely on external information via retrieval to access~\cite{kasai2023realtime,vu-etal-2024-freshllms}.
A similar ability has also been studied in the context of the forecasting task~\cite{zou2022forecasting,halawi2024approaching}.
While forecasting tasks predicting future outcomes based on historical information available \textit{prior} to an event, our real-time factuality evaluation setup allows access to \textit{contemporary} external evidence for cross-validation, which presents distinct challenges.

\section{Conclusion}
In this paper, we evaluate large language models on fake news detection of events that occur beyond their knowledge cutoff. We find that the conventional use of political claims from fact-checking websites is unsuitable for such tests due to emergent data shortcuts. We thus introduce an adversarial iterative pipeline to generate fake news that can gradually evade strong RAG-based detectors. Our empirical findings shed light on the behaviors of LLMs in both detecting and generating fake news about current world events. We hope that our framework and dataset will facilitate research efforts toward robust factual reasoning models under temporal distribution shift.
As malicious actors continue to refine their own generation techniques, our defenses must co-evolve, moving beyond static test sets towards dynamic environments that support continuously adaptive evaluation.

\section*{Limitations}

Our work focuses on evaluating prompting-based LLM detectors. LLM-generated fake news exhibits patterns that may differ from those found in human-written misinformation, which could limit the generalizability of our dataset for training detectors~\cite{zellers2019defending,huang-etal-2023-faking}. Future work could explore debiasing techniques—such as paraphrasing using the same LLM—to mitigate this issue~\cite{su2023fake}.

Our experiments primarily use English-language data centered on U.S. news, which may constrain the generalizability of our conclusions to other languages and regions. Expanding this research to multilingual and multicultural contexts is an important direction for future work. While our background study highlights common limitations and biases in popular fact-checking datasets, the findings are not intended to generalize to all fact-checking sources.

We use NBC News as a ground truth provider, assuming it to be a reliable and factual source. While we believe this is a reasonable assumption, no source is without bias. Importantly, our generation pipeline is adaptable and can be applied to alternative sources. Our fake news generation is grounded in real-world events, as opposed to fabricating entirely fictional scenarios, which may also spread harmful misinformation. However, during the rewriting process, the LLM may introduce hallucinated details not found in any external source. In such cases, the model must rely on its parametric knowledge to evaluate plausibility.

Finally, we note that directly evaluating the correlation between classifier performance on our dataset and real-world performance remains infeasible, as the distribution of misinformation is highly dynamic and shaped by ever-evolving events and generation strategies. Nonetheless, our dataset provides a necessary and reliable testbed—its factuality ensured through protocol design and human verification—for advancing the development and deployment of LLM-based detectors.

\section*{Ethics Considerations}
Our work aims to improve the robustness of fake news detection models by generating challenging fake news that can evade detection. We acknowledge the potential misuse of our method to create more deceptive misinformation. 
Simultaneously, we have shown that RAG-based detectors with high-quality retrieval can effectively counter such misinformation.
Our approach focuses on the factual reasoning aspect of fake news detection. Fake news generated does not usually contain propaganda, hate speech, or other harmful content.
The release of generators is critical to prepare detectors against adversarial attacks~\cite{zellers2019defending}. We responsibly released our code and data to facilitate further research on fake news detection.

\section*{Acknowledgements}
We thank Jun Yang, Junlin Wang and other members of the NLP group at Duke University for fruitful discussions.
This work was supported by NSF award IIS-2211526 and a gift from Together AI.

\bibliography{iclr2025_conference, anthology_0,anthology_1}

\appendix
\section{Appendix}

\begin{table}[h]
    \centering
    \begin{tabular}{ll}
        \toprule
        \textbf{Model Version} & \textbf{Knowledge Cutoff} \\
        \midrule
        gpt-3.5-turbo-0125 & September 2021 \\
        gpt-4-turbo-2024-04-09 & December 2023 \\
        gpt-4o-2024-05-13 & October 2023 \\
        gemini-1.5-pro-002 & November 2023 \\
        gemini-1.5-flash-002 & November 2023 \\
        Llama 3.1 (405B Instruct) & December 2023 \\
        \bottomrule
    \end{tabular}
    \caption{Versions and knowledge cutoff dates of the OpenAI, Gemini~\cite{team2024gemini}, and Llama~\cite{dubey2024llama} models used in our study.}
    \label{tab:knowledge_cutoffs}
\end{table}

\begin{table}[h]
    \centering
        \begin{tabularx}{\linewidth}{X}
            \toprule
            \textbf{2024-10-18} | As governor of Minnesota, Tim Walz ordered police to shoot residents with paintball guns for not obeying curfew orders during the COVID-19 pandemic.  \textbf{false} \\ \hline
            \textbf{2024-10-18} | Comedian and ``Family Feud'' game show host Steve Harvey died following a tragic accident in early 2024. \textbf{false}  \\ \hline
            \textbf{2024-10-19} | Research shows, for pet owners, the deaths of their animals can be just as hard as losing human loved ones. \textbf{true} \\ \bottomrule
        \end{tabularx}
    \caption{Examples of Snopes news from 2024.}
    \label{tab:politifact_samples}
\end{table}

\subsection{Snopes Experiment Details}
\label{sec:snopes}
We collect 3,212 Snopes news articles from January 2016 to October 2024. The data is accessed through the Google Fact Check Tools API.\footnote{\url{https://developers.google.com/fact-check/tools/api}}
These Snopes news are accompanied with true or false labels.

\begin{table}[h]
    \centering
        \begin{tabularx}{\linewidth}{X}
            \toprule
            \textbf{2024-08-07} | J.D. Vance says Tim Walz said he carried weapons in war, but “he has not spent a day in a combat zone.”  \textbf{true} \\ \hline
            \textbf{2024-08-09} | Elissa Slotkin says Mike Rogers “left Michigan to trade on his D.C. connections, helping Chinese tech companies get access to the U.S.” \textbf{false}  \\ \hline
            \textbf{2024-08-23} | Donald Trump: “Kamala cast the tiebreaking vote to hire 87,000 new IRS agents to go after your tip income.” \textbf{false} \\ \bottomrule
        \end{tabularx}
    \caption{Examples of PolitiFact news from 2024. Although all events happened after the knowledge cutoffs of LLMs, some of them can be reasonably classified using common sense.}
    \label{tab:politifact_samples}
\end{table}

\subsection{PolitiFact Experiment Details}
\label{sec:politifact}
\begin{figure}[htpb]
    \centering
    \includegraphics[width=\linewidth]{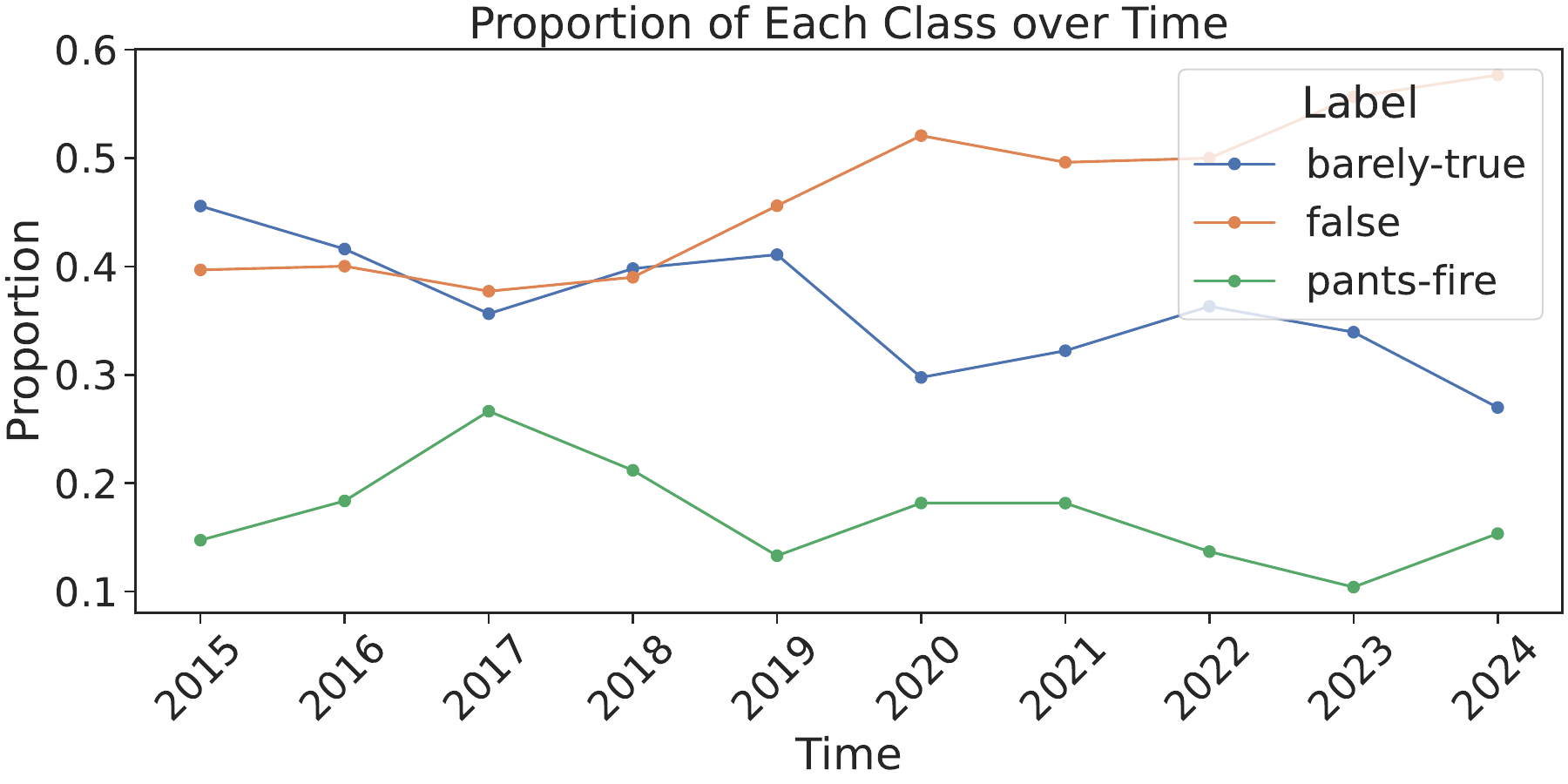}
    \caption{Proportion of three fake news classes of PolitiFact claims over the years, up to August 2024.}
    \label{fig:fake_prop_comparison}
\end{figure}

\begin{figure}[htpb]
    \centering
    \includegraphics[width=\linewidth]{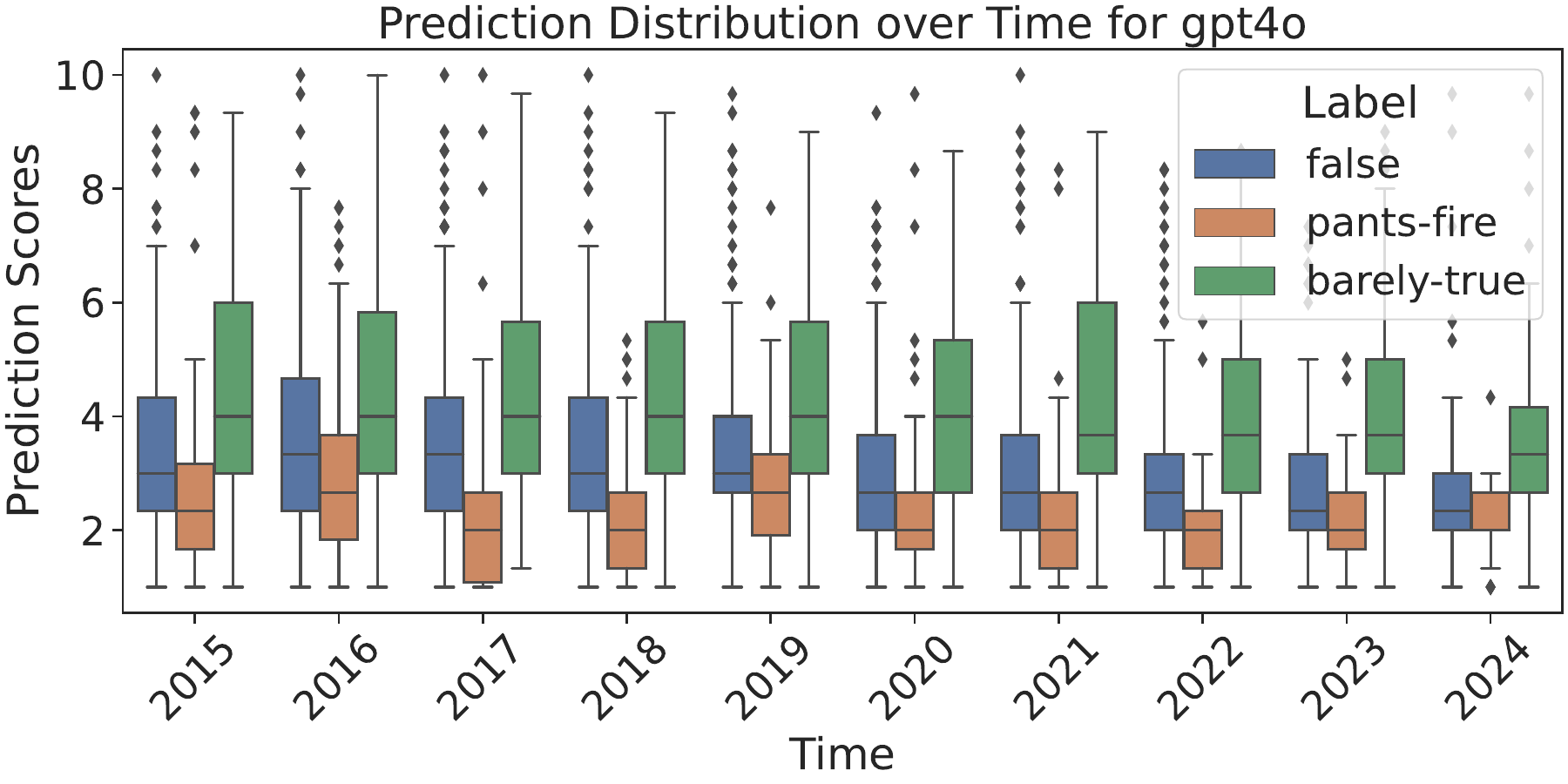}
    \caption{GPT-4o (retrieval-free) predicted plausibility of PolitiFact claims over the years, up to August 2024.}
    \label{fig:fake_dist_comparison}
\end{figure}

PolitiFact adopts six ﬁne-grained labels for the truthfulness ratings: pants-fire, false, barely-true, half-true, mostly-true, and true.
For the calculation of AUC-ROC, we adopt a common binarization to categorize the first three as negative and the last three as positive~\cite{liao2023muser,pelrine-etal-2023-towards}.
Digging into the increasing differential trend of the fake news classes, we find that it results from a combination of reasons: 1) there is an increasing proportion of the less truthful class in the data (i.e., `false' grows over `barely true' in \autoref{fig:fake_prop_comparison}); 2) both the `false' and `barely true' classes become less plausible (\autoref{fig:fake_dist_comparison}).

We verify the trend by fine-tuning a relatively small LM, RoBERTa~\cite{liu2019roberta}, on two sets of data from different years. In one experiment, we train the model on PolitiFact data from 2015 and evaluate it on data from 2016. In the other, we train the model on data from 2022 and evaluate it on data from 2023 and 2024. The results show that the later data results in a higher AUC-ROC score (0.765 versus 0.680), which is roughly equal to the performance of the GPT-3.5 detector.

\begin{figure*}[htpb]
    \centering
    \includegraphics[width=0.8\linewidth]{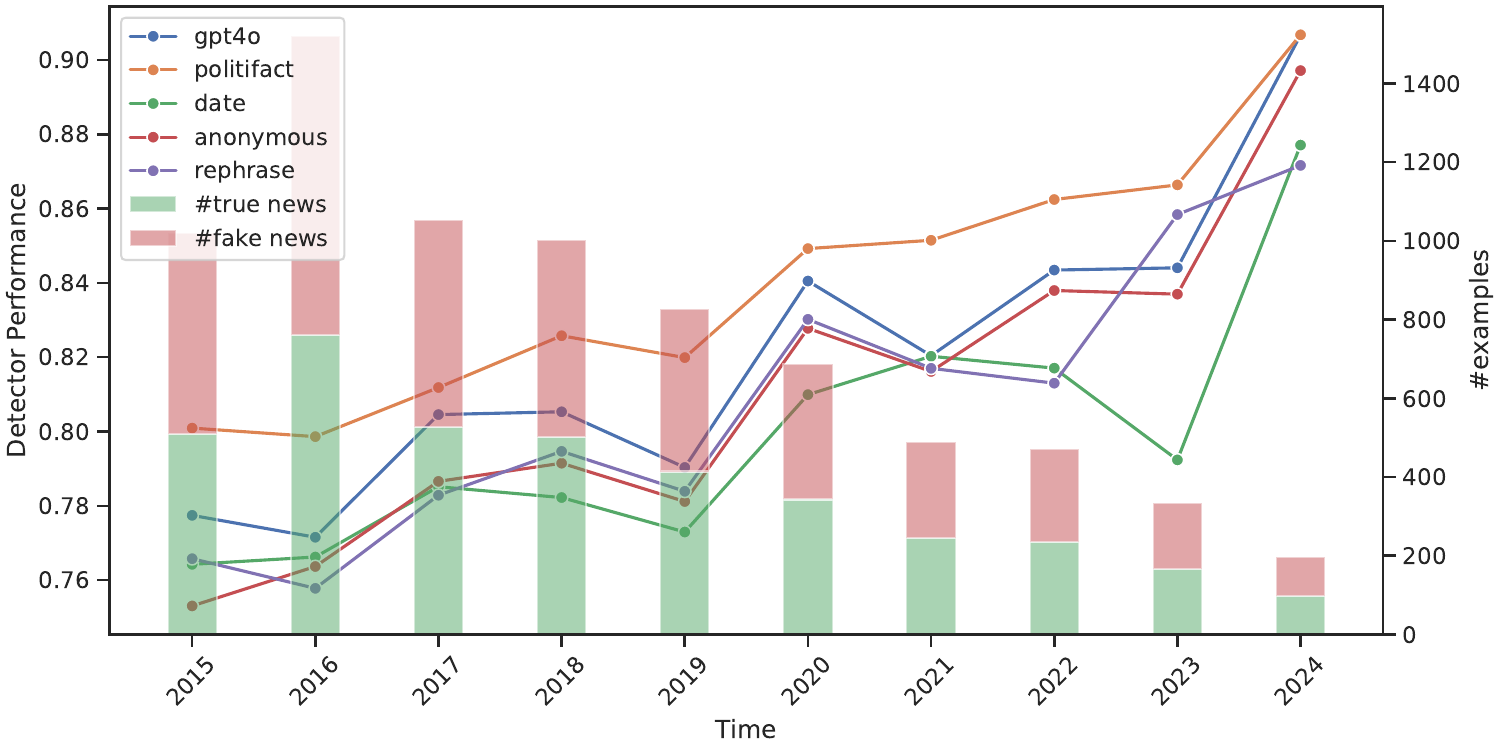}
    \caption{Ablating different attributes on PolitiFact claims over the years, up to August 2024. This figure shows the results on balanced data.}
    \label{fig:liar_time_ablation}
\end{figure*}

Ablating the attributes we use to classify a news story (\autoref{fig:liar_time_ablation}), we find that none of the attributes affect the increasing trend. Interestingly, we find that GPT-4o makes more accurate prediction when we ignore the date of publication in the prompt. Removing the originator of the claim is harmful, but not decisive—the classifier can judge the validity of a claim regardless of the speaker's credibility.

\begin{table}[h]
    \small
    \centering
\begin{tabular}{rrrr}
\toprule
\textbf{Year} & \textbf{\#Real} & \textbf{\#Fake} & \textbf{Real Prop.} \\
\midrule 2015 & 720 & 510 & 0.59 \\
 2016 & 938 & 758 & 0.55 \\
 2017 & 526 & 579 & 0.48 \\
 2018 & 501 & 625 & 0.44 \\
 2019 & 414 & 422 & 0.50 \\
 2020 & 343 & 578 & 0.37 \\
 2021 & 245 & 392 & 0.38 \\
 2022 & 236 & 402 & 0.37 \\
 2023 & 167 & 221 & 0.43 \\
 2024 & 97 & 189 & 0.34 \\
\bottomrule
\end{tabular}
\caption{The absolute numbers of both real and fake news selected in PolitiFact decrease, and the portion of real news in the range of [2020, 2024] decreases to an average of 38\% from an average of 51\% in the range of [2015, 2019]. 2024 data is up to August.}
\label{tab:politifact_prop}
\end{table}

Additionally, there is a meaningful distribution shift in PolitiFact data over the years. We observe that the proportion and the absolute number of true news has decreased significantly in recent years, which may result from the increasing prevalence of misinformation in the political landscape (\autoref{tab:politifact_prop}). From a machine learning perspective, this distribution shift leads to pronounced class imbalance, which can negatively impact the evaluation of fake news detection models if not properly accounted for.

\begin{table}[tpb]
    \centering
    \begin{tabular}{lrrrr} \toprule
& \textbf { AUROC } & \textbf { F1-Ma } & \textbf { Acc } \\
\midrule \text { Sampling } & 77.5 & 70.7 & 70.9 \\
 \text { 0 temp } & 76.2 & 69.4 & 69.4 \\ \bottomrule
    \end{tabular}
    \caption{GPT-4o detector with different sampling techniques on the LIAR dataset. Compared to taking one prediction with a temperature of 0, which is essentially the highest probability output, sampling with a temperature of 1 avoids overconfidence and produces smoother output.}
    \label{tab:detector_temp}
\end{table}

\begin{table}[tpb]
    \centering
    \begin{tabular}{lll}
    \toprule
 Event Range  & 2015-2018 & 2021-2024 \\ \midrule
 "Factuality" & 0.775 & 0.815 \\
 "Plausibility" & 0.753 & 0.822 \\ \bottomrule
\end{tabular}
    \caption{GPT-4o detector using prompts with different wording referring to the event likelihood. We report the AUROC on the PolitiFact data over the years. ``Factuality'' suits the fact-checking narrative of past events, while ``Plausibility'' better describes current and future events.}
    \label{tab:factual_vs_plausible}
\end{table}

\subsection{RAG Setup}
\label{sec:rag_setup}

To build an in-house RAG pipeline, we gather 811,000 news articles from various news sources within the same date range to create our retrieval corpus. We remove the exact seed true news from the retrieval corpus to avoid direct contamination. For popular news stories, multiple news portals may have similar coverage, which we allow for cross-verification. They still contain varying details that the detector can reason about (\autoref{tab:search_results2}).
A dense passage retriever~\cite{karpukhin-etal-2020-dense} pretrained on the Natural Questions  dataset~\cite{kwiatkowski-etal-2019-natural} is employed to index this corpus and provide 5 relevant articles when queried with a news headline.
We additionally implement an online RAG pipeline using SerpApi to retrieve 5 Google search snippets for the given news headlines.
We remove NBC News from the search results for similar reasons stated above.
Due to the cost of the API, we only use it for the final evaluation of the generated fake news.

\subsection{Details of Baseline Datasets}
\label{sec:baseline_dataset}
We compare to two recent LLM fake news datasets. We adopt the machine-written GossipCop and PolitiFact datasets from \citet{su2023fake}. The PolitiFact subset contains 132 true news and 97 fake news. The GossipCop dataset is down-sampled to 200 true and fake news. We adopt the \emph{rewrite} and \emph{information manipulation} variants of \citet{Chen24llm} generated by ChatGPT based on PolitiFact. The \emph{rewrite} subset contains 126 true news and 270 fake news. The \emph{information manipulation} subset contains 378 true news and 378 fake news from three sub-variants (i.e., Outdated Information, Description Ambiguity, False Context).

\subsection{Detector and Generator Prompts}
\label{sec:prompts}

We adopt simple prompts that implement corresponding functionality but do not introduce domain or dataset-specific heuristics to maximize generalizability.

\begin{tcolorbox}[breakable,colback=black!5!white, colframe=black!25!white, 
                  title=Detector, fonttitle=\bfseries, 
                  coltitle=black, sharp corners=south, 
                  boxrule=1pt, width=\linewidth]
\emph{
\texttt{<system>} Today is March 26, 2024. You predict the plausibility of a news you haven't seen. \texttt{</system>} \\ \\ 
Please predict the plausibility of the following news story: \\\texttt{<news/>} \\ 
Please give a number representing the plausibility (1-10) and nothing else.
}
\end{tcolorbox}

\begin{tcolorbox}[breakable,colback=black!5!white, colframe=black!25!white, 
                  title=RAG Detector, fonttitle=\bfseries, 
                  coltitle=black, sharp corners=south, 
                  boxrule=1pt, width=\linewidth]
\emph{
\texttt{<system>} Today is March 26, 2024. You predict the plausibility of a news you haven't seen given a list of related news stories from search results. \texttt{</system>} \\ \\ 
Related news stories from search results:\\\texttt{<retrieval/>} \\Please predict the plausibility of the following news story: \\\texttt{<news/>} \\ 
Please give a number representing the plausibility (1-10) and nothing else.
}
\end{tcolorbox}

\begin{tcolorbox}[breakable,colback=black!5!white, colframe=black!25!white, 
                  title=Detector Rationale, fonttitle=\bfseries, 
                  coltitle=black, sharp corners=south, 
                  boxrule=1pt, width=\linewidth]
\emph{
\texttt{<system>} Today is March 26, 2024. You fact-check a news you haven't seen, given a list of related news stories from search results. \texttt{</system>} \\ \\ 
Related news stories from search results:\\\texttt{<retrieval/>} \\Please predict the plausibility of the following news story: \\\texttt{<news/>} \\ 
Please explain what you find suspicious about the news story. Give the top 3 points in a list format and nothing else.
}
\end{tcolorbox}

\begin{tcolorbox}[breakable,colback=black!5!white, colframe=black!25!white, 
                  title=Generator, fonttitle=\bfseries, 
                  coltitle=black, sharp corners=south, 
                  boxrule=1pt, width=\linewidth]
\emph{
\texttt{<system>} Today is March 26, 2024. Please rewrite a real recent news story to introduce some believable factual errors. The modification should be able to fool professional fact-checkers. \\ Format\\Error introduced:\\Title:\\Description: \texttt{</system>} \\ 
\\Please rewrite the following news story. \\\texttt{<news/>} 
\\It was previously rewrote as: \\\texttt{<last\_iter\_news/>} \\ 
Fact-checkers found the following suspicious: \\\texttt{<rationale/>} \\ Please rewrite the news story to make it more believable and fool the fact-checkers.
}
\end{tcolorbox}

\begin{tcolorbox}[breakable,colback=black!5!white, colframe=black!25!white, 
                  title=Contradiction Detector, fonttitle=\bfseries, 
                  coltitle=black, sharp corners=south, 
                  boxrule=1pt, width=\linewidth]
\emph{
News 1\\\texttt{<news/>} \\ 
News 2\\\texttt{<fake\_news/>} \\ 
Does News 2 conflict with News 1? Please type yes or no and nothing else.
}
\end{tcolorbox}

\begin{table*}[t]
    \centering
    \begin{tabular}{lccccccccc}
    \toprule
    Retriever    & \multicolumn{3}{c}{None} & \multicolumn{3}{c}{News} & \multicolumn{3}{c}{Google} \\
    \cmidrule(lr){2-4} \cmidrule(lr){5-7} \cmidrule(lr){8-10}
    Detector       & first   & last  & $\Delta$  & first   & last  & $\Delta$  & first   & last   & $\Delta$   \\
    \midrule
    Gemini-Flash	&	54.6 & 50.6 & 4.0	&	73.9 & 62.4 & 11.5	&	80.0 & 72.2 & 7.7	\\
Gemini-Pro	&	56.3 & 52.1 & 4.3	&	69.7 & 59.2 & 10.5	&	74.9 & 66.0 & 8.8	\\
GPT-3.5	&	53.4 & 50.0 & 3.4	&	70.8 & 59.4 & 11.4	&	76.8 & 66.7 & 10.0	\\
GPT-4o	&	57.2 & 49.9 & 7.2	&	83.0 & 66.3 & 16.7	&	91.3 & 83.0 & 8.4	\\
Llama 3.1	&	57.4 & 52.3 & 5.1	&	80.9 & 69.2 & 11.7	&	91.3 & 83.2 & 8.2	\\
    \bottomrule
    \end{tabular}
    \caption{Average Precision (AP) of different detectors on the generated fake news of the first and last iteration. The $\Delta$ column shows the effects of the iterative process. \textit{News} refers to the in-house DPR retriever on news-please data, and \textit{Google} refers to the Google search API results.}
    \label{tab:main_results_pr}
    \end{table*}

\begin{table*}[thb]
    \footnotesize
    \centering
    \begin{tabularx}{\linewidth}{|X|}
    \hline
\textbf{Query:}    Saudi Arabia's parliamentary vote sees a low turnout despite government push
\\
    \hline
    \hline
\textbf{News (DPR)}\\
2024-03-11 - Government claims public’s lack of understanding of referenda led to landslide 'no' vote\\
Voters overwhelmingly rejected proposed changes to care the highest ever “no” vote percentage in an Irish referendum\\
\\
2024-03-05 - Opposition leaders react to the announcement of the date for the presidential elections\\
CARACAS After announcing the date for the next elections in Venezuela, guided by the CNE On July 28, leaders of the Venezuelan opposition expressed their\\
\\
2024-03-09 - Polls could have been derailed because of just one LHC order: CJP\\
Justice Faez underscores pivotal role outgoing Justice Tariq Masood could have played as LHC CJ but SC benefited immensely from his presence | Justice Tariq urges judges\\
\\
2024-03-04 - Senegal election crisis shakes support for Macky Sall’s coalition\\
DAKAR, March 4 (Reuters) – Writer Moustapha Gueye voted for Senegalese President Macky Sall at the last two elections. But disappointment in Sall’s second term and the president’s thwarted attempt to postpone the next vote have shaken Gueye’s allegiance to the ruling Benno Bokk Yakaar (BBY) coalition. Reclining on a sofa at his home in Dakar, Gueye […]\\
\\
2024-03-03 - The government's attempt for answers [Feb. 24-Mar. 1]\\
The federal government has ordered the executives of Bell, Rogers, and Telus to answer questions about telecom pricing after previous requests were denied.\\
\\\hline \hline
\textbf{Google} \\
Title: First Iranian parliament vote since 2022 mass protests sees ...\\
Source: PBS, Mar 1, 2024\\
Content: Iran held its first parliamentary election Friday since mass 2022 protests over mandatory hijab laws following the death of Mahsa Amini, apparently drawing a ...\\ \\
Title: Low turnout in Saudi Arabia's local polls | News\\
Source: Al Jazeera, Oct 22, 2011\\
Content: With women excluded until 2015, only men voted in kingdom's second-ever election, and polling booths remain mostly empty.\\ \\
Title: Growing 'Despondency' And Hard-Liners' Dominance\\
Source: Radio Free Europe/Radio Liberty, \\
Content: Iranian President Ebrahim Raisi casts his vote during parliamentary elections in Tehran on March 1. “The Islamic republic is now a minority-ruled ...\\ \\
Title: Democracy in Crisis\\
Source: Freedom House, \\
Content: Political rights and civil liberties around the world deteriorated to their lowest point in more than a decade in 2017, extending a period characterized by ...\\ \\
Title: In Saudi Arabia, Only Men Vote, And Not Often\\
Source: NPR, Sep 29, 2011\\
Content: Only men could vote in polls to fill half the seats on some 300 municipal councils. The other half are appointed by the government." \\ \hline
    \end{tabularx}
    \caption{Sample retrieval results corresponding to example 3 (round 1) in Table~\ref{tab:examples}, the search query is about ``Saudi Arabia'', Google robustly returns ``Iran'' results.}
    \label{tab:search_results1}
\end{table*}

\begin{table*}[thb]
    \footnotesize
    \centering
    \begin{tabularx}{\linewidth}{|X|}
    \hline
\textbf{Query:}    Iranian parliamentary vote sees a low turnout despite government push
\\
    \hline
    \hline
\textbf{News (DPR)}\\
"Related news stories from search results:\\
\\
2024-03-01 - Iranian parliament vote, first since 2022 mass protests, sees a low turnout despite government push\\
Iran has held its first parliamentary election since mass 2022 protests over mandatory hijab laws after the death of Mahsa Amini, apparently drawing a low turnout amid calls for a boycott. It wasn’t immediately clear if voter apathy or an active desire to send a message to Iran’s theocracy depressed the number of voters coming to polling stations Friday across the Islamic Republic. While state-controlled television broadcast images of lines of voters, others across the capital of Tehran saw largely empty polling stations. Some, including imprisoned Nobel Peace Prize laureate Narges Mohammadi, urged a boycott of a vote they derided as a “sham.""\\
\\
2024-03-01 - Iranian Parliament Vote, First Since 2022 Mass Protests, Sees a Low Turnout Despite Government Push\\
Get latest articles and stories on World at LatestLY. Iran held its first parliamentary election on Friday since mass 2022 protests over mandatory hijab laws following the death of Mahsa Amini, apparently drawing a low turnout amid calls for a boycott.\\
World News | Iranian Parliament Vote, First Since 2022 Mass Protests, Sees a Low Turnout Despite Government Push.\\
\\
2024-03-01 - Iranian parliament vote, first since 2022 mass protests, sees low turnout despite government push\\
It wasn’t immediately clear if voter apathy or an active desire to send a message to Iran’s theocracy depressed the number of voters coming to polling stations across the Islamic Republic.\\
\\
2024-03-11 - Government claims public’s lack of understanding of referenda led to landslide 'no' vote\\
Voters overwhelmingly rejected proposed changes to care the highest ever “no” vote percentage in an Irish referendum\\
\\
2024-03-02 - Low turnout in Iran's first vote since 2022 protests\\
Iran's voters have been reluctant to turn out in the country's first parliamentary election since protests over the death in custody of Mahsa Amini in 2022.\\
\\ \hline \hline
\textbf{Google} \\
Title: First Iranian parliament vote since 2022 mass protests sees ...\\
Source: PBS, Mar 1, 2024\\
Content: Iran held its first parliamentary election Friday since mass 2022 protests over mandatory hijab laws following the death of Mahsa Amini, apparently drawing a ...\\ \\
Title: Iranian parliament vote, first since 2022 mass protests ...\\
Source: Euronews.com, Mar 1, 2024\\
Content: Officials including Supreme Leader Ayatollah Ali Khamenei sought to link turnout directly to taking a stand against Iran's enemies.\\ \\
Title: Hard-liners dominate Iran parliamentary vote that saw a ...\\
Source: AP News, Mar 4, 2024\\
Content: Iranian hard-line politicians dominated the country's vote for parliament. However, the election Friday also saw a record-low turnout.\\ \\
Title: Low turnout as conservatives dominate Iran parliamentary ...\\
Source: Al Jazeera, Mar 4, 2024\\
Content: Conservative politicians will dominate Iran's parliament, according to election results, maintaining their hold on the Islamic Consultative ...\\ \\
Title: Iranian parliament vote sees low turnout\\
Source: The Sydney Morning Herald, Mar 2, 2024\\
Content: Iran on Friday held its first parliamentary election since mass 2022 protests over mandatory hijab laws following the death of Mahsa Amini. \\ \hline
    \end{tabularx}
    \caption{Sample retrieval results corresponding to example 3 (round 3) in Table~\ref{tab:examples}, the search query is about ``Iran'', both search engines return relevant results for cross-verification.}
    \label{tab:search_results2}
\end{table*}

\begin{figure*}[htpb]
    \centering
    \includegraphics[width=0.8\linewidth]{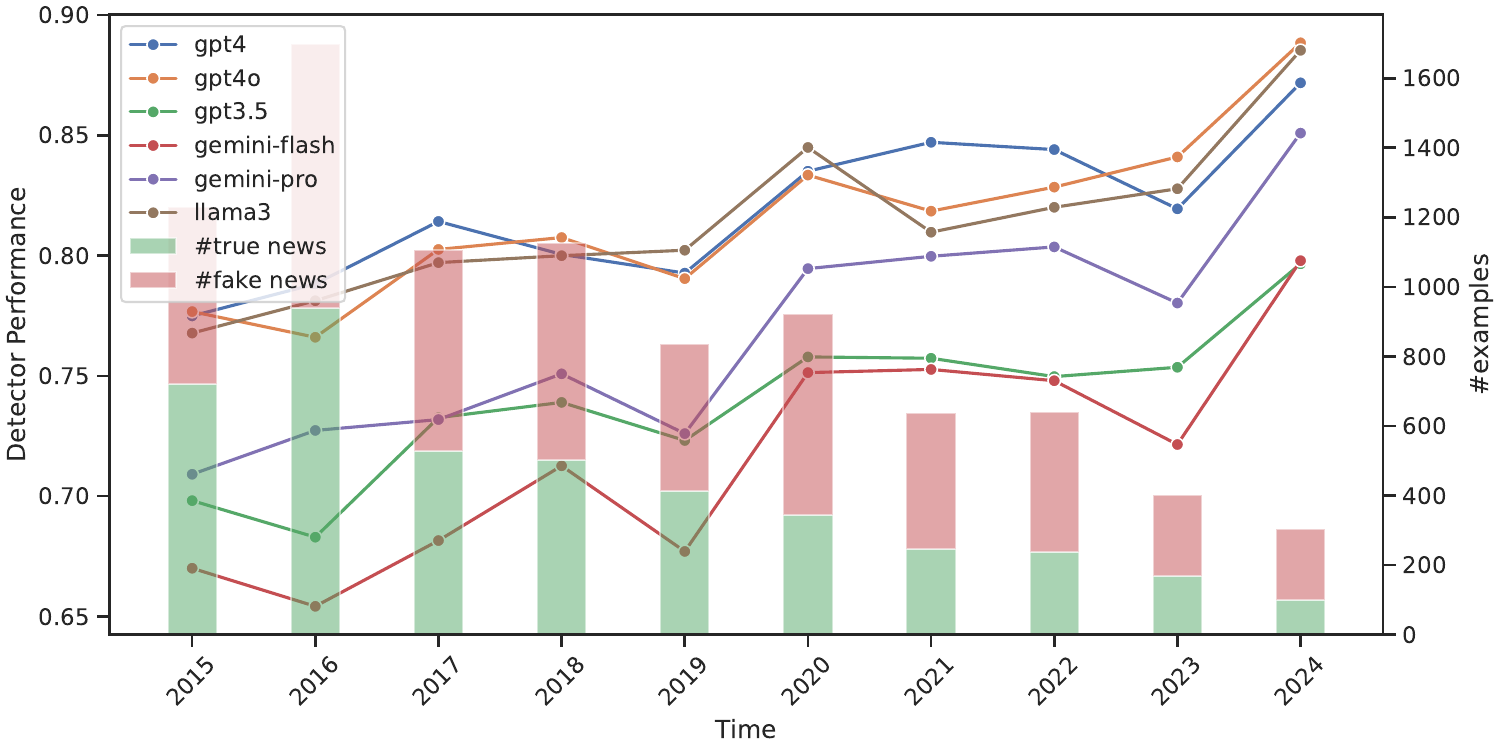}
    \caption{Comparison of different detectors (AUC-ROC) on PolitiFact claims over the years, up to August 2024. This figure shows the results on raw data without balancing. The absolute number and proportion of true news decrease throughout the year.}
    \label{fig:auc_comparison_raw}
\end{figure*}

\begin{figure}[htpb]
    \centering
    \includegraphics[width=\linewidth]{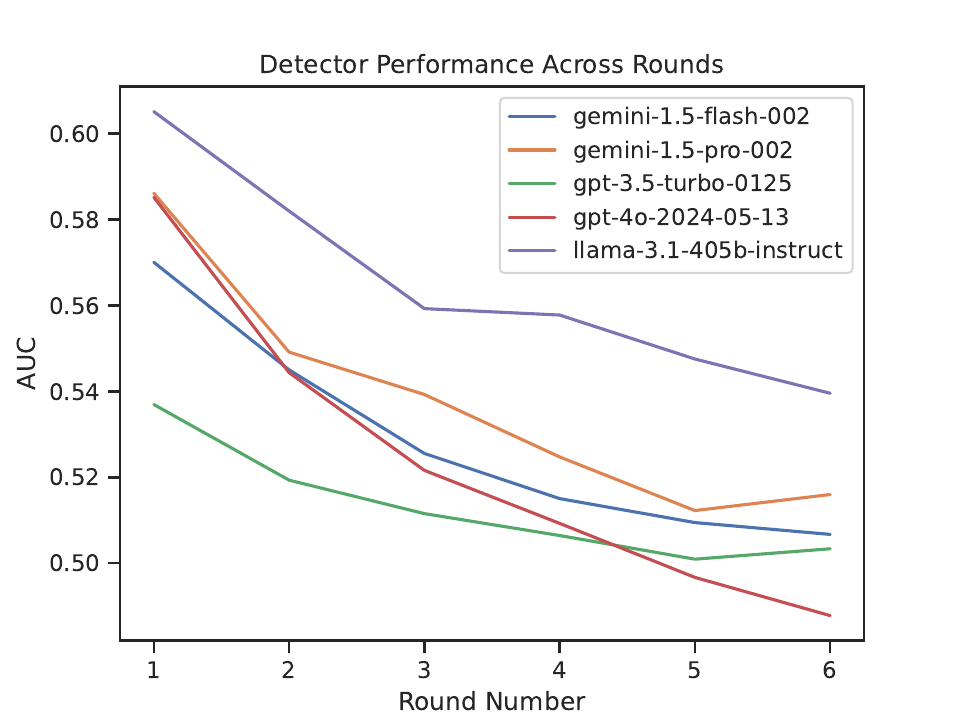}
    \caption{Non-RAG-based detectors with different LLM backbones (AUC-ROC) on iteratively generated fake news.}
    \label{fig:direct_rounds}
\end{figure}

\begin{figure}[htpb]
    \centering
    \includegraphics[width=\linewidth]{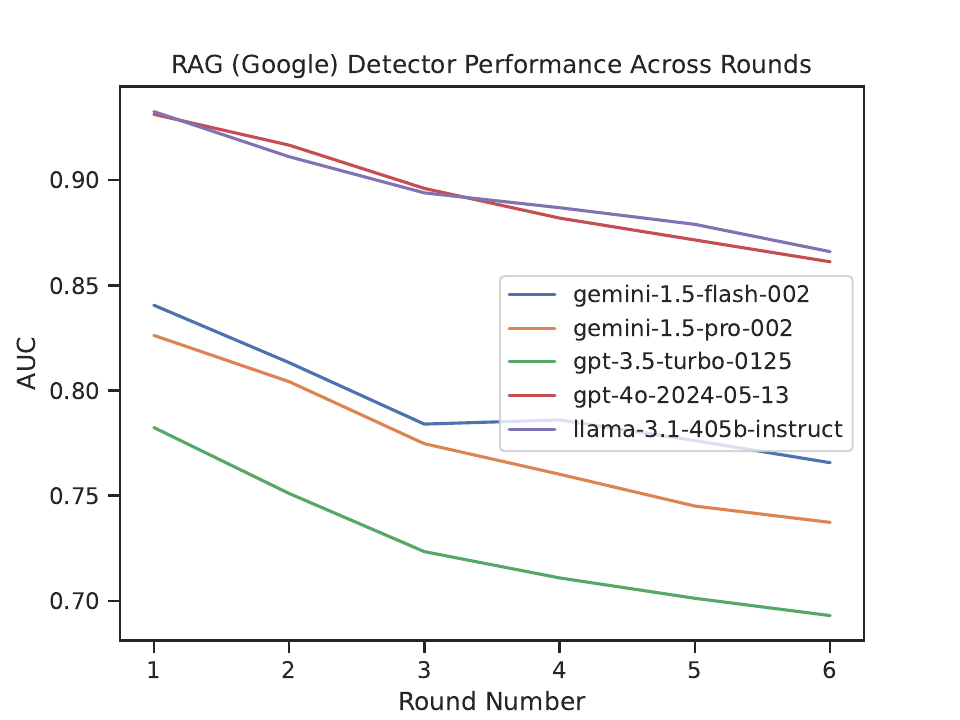}
    \caption{Google-RAG-based detectors with different LLM backbones (AUC-ROC) on iteratively generated fake news.}
    \label{fig:google_rounds}
\end{figure}

\begin{table*}[t]
    \small
    \centering
    \begin{tabularx}{\linewidth}{llXlr}
    \toprule
    ID & Originator & Claim & Date \\ \midrule
    11593 & Marco Rubio & Says Ted Cruz "is a supporter of legalizing people that are in this country illegally" and "proposed giving them work permits." & November 12, 2015 \\ \hline
11099 & Tom Cotton & President Barack Obama "said at the beginning of the negotiations that the basic approach was to dismantle Iran’s nuclear program in exchange for dismantling the sanctions." & July 15, 2015 \\ \hline
11206 & Bernie Sanders & "We spend almost twice as much per capita on health care as do the people of any other country." & August 16, 2015 \\ \hline
10672 & Sally Kohn & "White men account for 69 percent of those arrested for violent crimes." & March 19, 2015 \\ \hline
10845 & Alex McMurtrie Jr. & State legislators "quietly shifted \$2 billion from education to road building" in 2013. & May 7, 2015 \\ \hline
11741 & Steven Landes & Medicaid expansion "could cost the Commonwealth of Virginia over \$1 billion a year..." & December 24, 2015 \\ \hline
11443 & Steven Costantino & "I did not play any role in bringing the company to RI as did others in government. I was tasked with handling the legislation affecting the company by my superiors." & September 27, 2015 \\ \hline
10738 & Judicial Watch & "ISIS camp a few miles from Texas, Mexican authorities confirm." & April 14, 2015 \\ \hline
10680 & Martin Smith & By allowing brewpubs to sell beer, Georgia could become like Mexico with only a couple of manufacturers controlling all aspects of market. & March 23, 2015 \\ \hline
11282 & Ted Cruz & "The Iran Deal will facilitate and accelerate the nation of Iran acquiring nuclear weapons." & September 9, 2015 \\ \midrule
    \end{tabularx}
    \caption{PolitiFact collected fake news from 2015, randomly sampled from examples that are assigned a ``false'' Truth-O-Meter label.}
\end{table*}

\begin{table*}[t]
    \small
    \centering
    \begin{tabularx}{\linewidth}{llXlr}
    \toprule
    ID & Originator & Claim &  Date \\ \midrule
    25742 & Brigitte Gabriel & "For 18 months under President Trump, not a single American was harmed in Afghanistan.” & July 2, 2024 \\ \hline
25078 & Steve Scalise & The Senate’s border bill “accepts 5,000 illegal immigrants a day.” & February 4, 2024 \\ \hline
25020 & Ron DeSantis & Says Winston Churchill said, “Success is not final, failure is not fatal: it is the courage to continue that counts.” & January 21, 2024 \\ \hline
25800 & Jon Stewart & Milwaukee's Marcus Performing Arts Center – where 'The Daily Show' had been scheduled – “was originally located in the ‘soft perimeter,’ they called it, security-wise” but “was shifted, understandably so, to the ‘hard perimeter.’” & July 16, 2024 \\ \hline
25553 & Ron Johnson & "Every Senate Democrat has voted to support unlimited abortions up to the moment of birth.” & April 15, 2024 \\ \hline
25596 & Jesse Watters & Judge Juan Merchan “overrules every objection from the defense and sustains every objection from the prosecution” during former President Donald Trump’s New York trial. & May 28, 2024 \\ \hline
25240 & Donald Trump & “Biden has implemented a formal policy that illegal aliens who intrude into the United States are granted immunity from deportation.” & March 9, 2024 \\ \hline
25376 & Donald Trump & "Crime is down in Venezuela by 67\% because they're taking their gangs and their criminals and depositing them very nicely into the United States.” & April 2, 2024 \\ \hline
25956 & Eric Hovde & Says U.S. Sen. Tammy Baldwin “has done absolutely nothing” about the fentanyl crisis. & August 13, 2024 \\ \hline
25082 & Elon Musk & Biden’s strategy is to “get as many illegals in the country as possible” and “legalize them to create a permanent majority.” & February 2, 2024 \\ \midrule
    \end{tabularx}
    \caption{PolitiFact collected fake news from 2024, randomly sampled from examples that are assigned a ``false'' Truth-O-Meter label.}
\end{table*}

\end{document}